\documentclass[sn-mathphys]{sn-jnl}
\usepackage{color}
\usepackage{colortbl} 
\usepackage{xcolor}
\usepackage{pifont}
\usepackage{graphicx}
\usepackage{multicol}
\usepackage{multirow}
\usepackage{array}
\usepackage{xspace}
\usepackage{enumitem}
\setlist[itemize]{leftmargin=*}
\setlist[enumerate]{leftmargin=*}

\usepackage{amsmath}
\usepackage[font=small,labelfont=bf]{caption}
\usepackage{tablefootnote}
\usepackage{listings}
\usepackage{algorithm}
\usepackage{algorithmicx}

\usepackage{relsize}

\definecolor{codegreen}{rgb}{0,0.6,0}
\definecolor{codegray}{rgb}{0.5,0.5,0.5}
\definecolor{codepurple}{rgb}{0.58,0,0.82}
\definecolor{backcolour}{rgb}{0.95,0.95,0.92}
\lstdefinestyle{mystyle}{
    frame=single,
    framesep=12pt,
    backgroundcolor=\color{backcolour},   
    commentstyle=\color{codegreen},
    keywordstyle=\color{magenta},
    numberstyle=\tiny\color{codegray},
    stringstyle=\color{codepurple},
    basicstyle=\ttfamily\footnotesize,
    breakatwhitespace=false,         
    breaklines=true,                 
    captionpos=b,                    
    keepspaces=true,                 
    numbers=left,                    
    numbersep=5pt,                  
    showspaces=false,                
    showstringspaces=false,
    showtabs=false,                  
    tabsize=2,
    xleftmargin=.025\textwidth, 
    xrightmargin=.025\textwidth
}

\jyear{2023}%

\theoremstyle{thmstyleone}%
\theoremstyle{thmstyletwo}%

\theoremstyle{thmstylethree}%

\newcommand{\ours}{\text{Open-TI}\xspace}

\begin{document}

\title[Open Traffic Intelligence with Augmented Language Model]{Open-TI: Open Traffic Intelligence with Augmented Language Model}










\author[1]{\fnm{Longchao} \sur{Da}}
\email{longchao@asu.edu}

\author[1]{\fnm{Kuanru} \sur{Liou}}\email{kliou@asu.edu}

\author[1]{\fnm{Tiejin} \sur{Chen}}\email{tchen169@asu.edu}


\author[2]{\fnm{Xuesong} \sur{Zhou}}\email{xzhou74@asu.edu}

\author[2]{\fnm{Xiangyong} \sur{Luo}}\email{xluo70@asu.edu}

\author[1]{\fnm{Yezhou} \sur{Yang}}\email{yz.yang@asu.edu}

\author[1]{\fnm{Hua} \sur{Wei}}\email{hua.wei@asu.edu}
\equalcont{Corresponding Author.}

\affil*[1]{\orgdiv{School of Computing and Augmented Intelligence}, \orgname{Arizona State University}, 
\orgaddress{\street{350 E Lemon St}, \city{Tempe}, \postcode{85287}, \state{AZ}, \country{USA}}}

\affil[2]{\orgdiv{School of Sustainable Engineering and the Built Environment}, \orgname{Arizona State University}, \orgaddress{\street{350 E Lemon St}, \city{Tempe}, \postcode{85287}, \state{AZ}, \country{USA}}}


\abstract{
Transportation has greatly benefited the cities' development in the modern civilization process. Intelligent transportation, leveraging advanced computer algorithms, could further increase people's daily commuting efficiency. However, intelligent transportation, as a cross-discipline, often requires practitioners to comprehend complicated algorithms and obscure neural networks, bringing a challenge for the advanced techniques to be trusted and deployed in practical industries. Recognizing the expressiveness of the pre-trained large language models, especially the potential of being augmented with abilities to understand and execute intricate commands, we introduce Open-TI. Serving as a bridge to mitigate the industry-academic gap, Open-TI is an innovative model targeting the goal of \textbf{T}uring \textbf{I}ndistinguishable \textbf{T}raffic \textbf{I}ntelligence, it is augmented with the capability to harness external traffic analysis packages based on existing conversations. Marking its distinction, Open-TI is the first method capable of conducting exhaustive traffic analysis from scratch  - spanning from map data acquisition to the eventual execution in complex simulations. Besides, Open-TI is able to conduct task-specific embodiment like training and adapting the traffic signal control policies (TSC), explore demand optimizations, etc. Furthermore, we explored the viability of LLMs directly serving as control agents, by understanding the expected intentions from Open-TI, we designed an agent-to-agent communication mode to support  Open-TI conveying messages to ChatZero (control agent), and then the control agent would choose from the action space to proceed the execution. We eventually provide the formal implementation structure, and the open-ended design invites further community-driven enhancements. }

\keywords{Large Language Models, Traffic Simulation, Traffic Signal Control}



\maketitle

\section{Introduction}\label{sec1}
Traffic and Transportation have played an important role in the process of human civilization. With the development of electronic and computer techniques, intelligent transportation is casting hope to further benefit people's daily lives through optimal controlling and scheduling decisions. Transportation depicts vehicles commuting between regions, providing delivery for products and humans. The efficient modern transportation comes from joint efforts from many researchers in various directions like: map modeling~\cite{yukawa1995coupled}, traffic simulation~\cite{chao2020survey}, schedule optimization~\cite{dai2020joint}, etc., and still, there are multiple ongoing challenges regarding the multi-resolution traffic simulation~\cite{zhou2022meso}, optimal traffic signal control policies~\cite{wei2019colight}, dynamic demand dispatch adjustment~\cite{osorio2019high}, etc. More specifically, when it comes to vehicle control, the intelligent traffic signal brings hope to city-scale congestion mitigation and energy saving, multiple frontier solutions have been released on different simulators, such as SUMO~\cite{lopez2018microscopic}, CityFlow~\cite{zhang2019cityflow}, VISSIM~\cite{fellendorf2010microscopic}. These simulators and algorithms are powerful, and efficient, but hard to operate and implement, thus, introducing a gap from the research to the industry, and leading to a trustworthy problem for practitioners.

The possible solution to bridge that gap includes two steps: 1. Unifying the simulation and analysis process by a standard ecosystem like General Modeling Network Specification (GMNS)~\cite{lu2023virtual} to define a common format for sharing routable road network files and is designed for multi-modal static and dynamic transportation planning and operations. 2. Building an intelligent system with self-explain abilities, which is integrated with multiple domain-specific tasks and the corresponding frontier solutions: state-of-the-art algorithms, powerful simulators, etc., and can be easily executed with sufficient explanations in an interactive way. 

\begin{figure}[h!]
    \centering
    \includegraphics[width=0.75\linewidth]{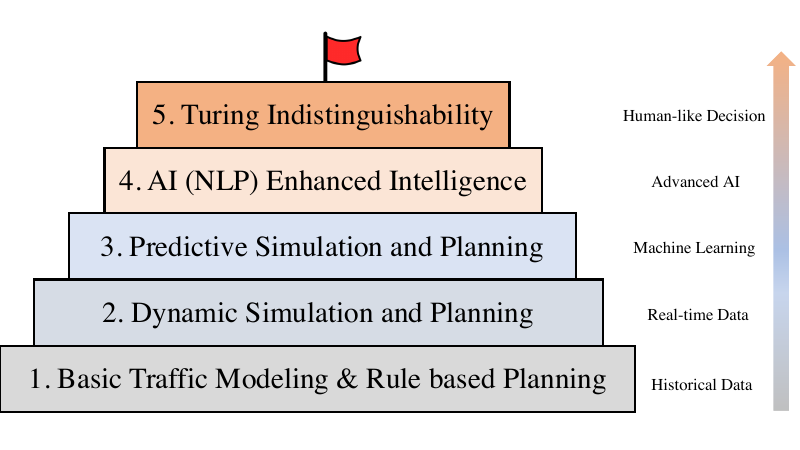}
    \caption{The 5 Stages of Development of Traffic Intelligence}
    \label{fig:levels}
\end{figure}

Transportation intelligence can be divided into 5 stages according to the development of technology as in Fig.~\ref{fig:levels}, with Turing Indistinguishable as the hardest one to provide human-like decisions. 
With large language models becoming increasingly instrumental in aiding humans, many researchers have leveraged the LLMs to benefit transportation tasks~\cite{zhang2023trafficgpt, de2023llm} in stage 4, and this has provided a clearer vision on stage 5 as a more Turing Indistinguishable traffic intelligence. Large-scale pre-trained models such as Llama, GPT-3.0, and ChatGPT, are endowed with the capacity to grasp the context, dissect issues, and discern the logic connecting questions with answers, which can deliver in-depth clarifications on specific topics through a sequence of interactive dialogues. Early explorations are made by leveraging LLMs to benefit domain-specific tasks, such as: Copilot~\cite{vaithilingam2022expectation}, DrugGPT~\cite{li2023druggpt}, TrafficGPT~\cite{zhang2023trafficgpt}, GraphGPT~\cite{tang2023graphgpt}, etc. Due to the limitation of only tackling the context-level questions, researchers managed to augment the language model on their ability to take action and use tools, which significantly broadened the application scenarios and enhanced the beneficial impact~\cite{mialon2023augmented}. Defined as \textbf{Augmented Language Models (ALMs)}, this refers to `language models (LMs) that are augmented with reasoning skills and the ability to use tools'~\cite{mialon2023augmented}. Inspired by ALMs, we propose to design a rudiment of Turing Indistinguishable Traffic Intelligence: Open-TI, an augmented traffic agent not only able to provide conversational insights, but also able to understand human intentions, conduct intelligent traffic analysis from scratch, answer questions regarding the used techniques or tools, and provide an interpretation of the results. By this, it will be more convenient for the industrial practitioners or any stakeholders to learn about traffic and transportation studies, and cast interesting case analyses.

In this paper, we first provide a background introduction to recent most related research directions, including traffic simulations, traffic domain-specific tasks, and augmented language models. Then we propose a pivotal augmented language agent \ours, that is integrated with a neat interface to operate possible tools, thus realizing the language-level operation, it is worth noting that, the \ours is able to conduct traffic analysis from scratch (from downloading map to provide simulation on the interested area). Second, we realized multiple domain-specific tasks like traffic signal control policy training and traffic demand optimization under the unified implementation class. Third, we design the ChatLight to realize the meta-control based on the LLM's inference ability, by agent-agent communication, the pivotal agent will interact with humans to obtain control requirements and convey the message to the ChatZero control agent, then, the ChatZero will conduct the decision making based jointly on the observation of current traffic situation and described policies.\footnote{We have released the code at:\url{https://github.com/DaRL-LibSignal/OpenTI.git}.}

In summary, the \textbf{contributions} of this paper are:

\begin{itemize}
    \item \ours is the first paradigm to realize a thorough and stable traffic analysis from scratch: from downloading maps to simulation of interested areas.
    \item \ours is a central agent in charge of multiple task-specific executors, including traffic signal control, demand optimization, etc. It is equipped with standard API for open implementation access of the research community to enhance its capabilities.
    \item \ours could realize meta-control: convey the policy description by communicating with another control agent that directly serves as a controller to output actions based on context understanding. 
    \item \ours provides sufficient explanations to any questions during the traffic analysis, from log file interpretation to results comparison, bringing convenience for traffic management and transportation strategy planning. 
\end{itemize}

\section{Background and Related Work}\label{sec2}

This section provides concepts of augmented language agents, traffic simulation, and transportation research tasks.

\subsection{Augmented Language Models}
Large Language Models (LLMs)~\cite{devlin2018bert, brown2020language, chowdhery2023palm} have boosted dramatic progress in Natural Language Processing (NLP) and are already core in several products with millions of users, such as the coding assistant Copilot~\cite{chen2021evaluating}, Bert enhanced search engine\footnote{\url{https://blog.google/products/search/search-language-understanding-bert/}} and  2. ChatGPT and GPT4~\cite{liu2023summary}. LLMs are able to execute multiple tasks from language understanding to conditional and unconditional text generation relying on memorization~\cite{tirumala2022memorization} combined with compositionality~\cite{zhou2022least} capabilities, thus opening a realistic path towards higher-bandwidth human-computer interactions, or even benefit other research domains by its inference ability~\cite{da2023llm}. But LLMs are not held solely in the text conversation, when LLMs are equipped with the tools using abilities, it will bring more changes to people's lives. Some literature shows that by augmenting the LLMs with the tool-using ability, it could realize the advanced automation and bring the intelligence science closer to the goal of being Turing Indistinguishable, such as~\cite{li2023api} design the API bank to execute API calls to meet human needs,~\cite{wang2023augmenting} applied the augmented language models to the medical field to serve as a more flexible knowledge hub for doctors and patients.~\cite{liang2023taskmatrix} focuses more on using existing foundation models (as a brain-like central system) and APIs of other AI models and systems (as sub-task solvers) to achieve diversified tasks in both digital and physical domains. Our work is the first to explore the augmented language agents on automatic traffic intelligence that realize a throughout traffic analysis system.


\subsection{Traffic Simulation}
\begin{figure}[h!]
    \centering
    \includegraphics[width=0.95\linewidth]{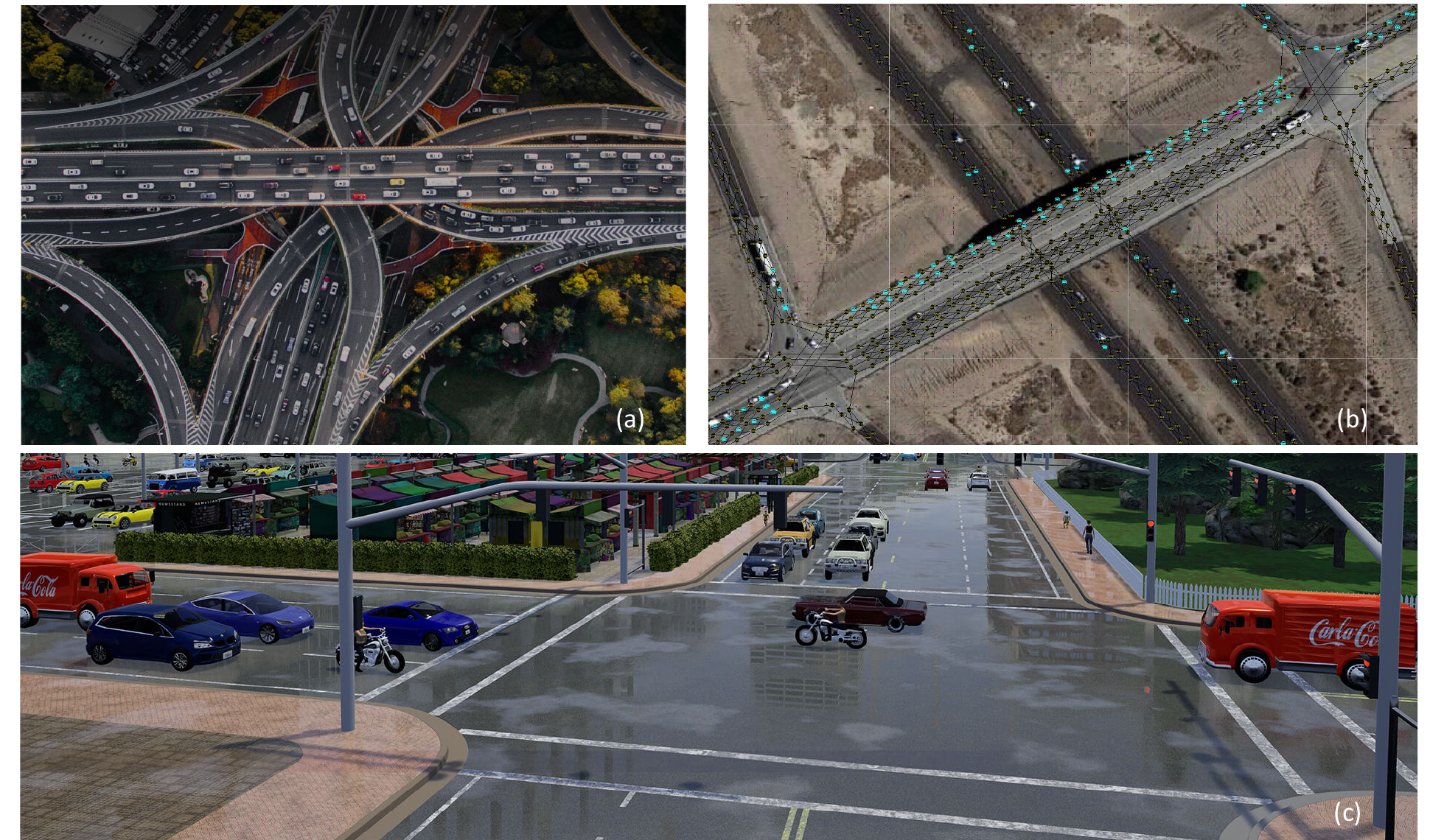}
    \caption{The traffic and transportation simulation in cities, (a) is a real-world traffic image, (b) is the simulation of traffic flow in DTALite~\cite{tong2019open}, and (c) is the city simulation from CARLA~\cite{dosovitskiy2017carla}.}
    \label{fig:simulation}
\end{figure}

The continuous growth of urban populations and the increase in vehicular traffic have accentuated the need for efficient traffic management and planning. Traffic simulation provides a good reference for planning strategies, offering insights into traffic patterns, road network efficiencies, and the potential impacts of infrastructural changes as shown in Fig.~\ref{fig:simulation}. The utilization of traffic simulation models facilitates the analysis of traffic behavior under various conditions without the need for costly real-world alterations. 

Traffic simulation has gone through various stages, from simplistic models to complex systems incorporating real-time data and predictive analytics. Early efforts in traffic simulation were concentrated on microscopic models, which simulate individual vehicle movements. Pioneering work by~\cite{mullakkal2020hybrid} and others laid the foundation for microscopic traffic simulation, focusing on the behavioral aspects of individual drivers. As computing power increased, mesoscopic and macroscopic models gained popularity. Mesoscopic models, like those developed by~\cite{de2019mesoscopic}, provide a balance between simulating details and computational efficiency, ideal for medium-scale traffic networks. Macroscopic models, on the other hand, such as the work by~\cite{oppe1989macroscopic}, offer insights into broader traffic flow trends, suitable for large-scale analysis but lacking the granularity of microscopic models.

Recent advancements have shifted focus towards data-driven and AI-based approaches. The integration of machine learning, as seen in the work of~\cite{boukerche2020artificial}, has enabled more accurate predictions of traffic patterns by learning from historical data. Additionally, the incorporation of real-time data, such as that from IoT devices and traffic sensors~\cite{masek2016harmonized}, has significantly enhanced the responsiveness of traffic simulation models. Studies by~\cite{maroto2006real} have demonstrated the effectiveness of real-time data in adapting traffic simulations to dynamic conditions. 

With the development of autonomous vehicles (AVs), traffic simulation for safety tests and validation become important. The simulator developed by NVIDIA~\cite{nvidiasim} tends to provide a more physically accurate simulation platform and Waymax~\cite{gulino2023waymax}, which is developed by autonomous vehicle company Waymo, provides a multi-agent scene simulator on hardware accelerators to empower the simulation for AVs.

In this work, we provide support to multiple simulators like SUMO~\cite{behrisch2011sumo}, CityFlow~\cite{zhang2019cityflow}, and DLSim~\cite{tong2019open}, and further present open-implementation instruction to help integrate more advanced ones with the development of the research community.

\subsection{Traffic Domain Specific Tasks}

\paragraph{Traffic Signal Control}
Traffic Signal Control (TSC) is crucial for improving traffic flow, reducing congestion in modern transportation systems, and benefiting individuals and societies. Traffic signal control remains an active research topic because of the high complexity of the problem. The traffic situations are highly dynamic and require traffic signal plans to adapt to different situations, making it necessary to develop effective algorithms that can adjust to changing traffic conditions~\cite{qadri2020state}. Recent advances in reinforcement learning (RL) techniques have shown superiority over traditional approaches in TSC~\cite{wei2018intellilight}. In RL, an agent aims to learn a policy through trial and error by interacting with an environment to maximize the cumulative expected reward over time, when learning the policy, it takes the features like intersection vehicle amount, and current traffic light phase (state) as observations, and change the traffic light phases or duration times (actions), to realize the relief of traffic congestion (reward). 
The most evident advantage of RL is that it can directly learn how to generate adaptive signal plans by observing the feedback from the environment, so it is beneficial for traffic intelligence to integrate TSC tasks to provide planning insights.


\paragraph{OD Matrix Optimization}
Origin-destination (O-D) matrix estimation is a critical component in the field of transportation planning and traffic engineering and involves the creation of a matrix that represents the number of trips between various origins and destinations in a given area over a specified time period. This matrix is fundamental for understanding travel demand patterns and is important for traffic modeling, infrastructure planning, and policy-making in urban environments. The process of O-D matrix estimation can vary in complexity depending on the size of the area being studied and the granularity of the data required. Traditional methods often involve the use of travel surveys, where individuals report their travel behavior, or the extrapolation of data from limited traffic counts~\cite{Willumsen1978, Abrahamsson1998, Medina2002}. However, these methods can be time-consuming and may not always capture the full spectrum of travel patterns due to their reliance on sampling or incomplete data.

The O-D matrix estimation has evolved in recent years, modern techniques leverage large datasets obtained from a variety of sources, such as traffic sensors, GPS devices, and mobile phone signals~\cite{Mahmassani2001, Mahmassani2005, Zhou2003, Zhou2006, Zhou2010, Zhou2013,Krishnakumari2020,Fedorov2019, Pamu2023, Fu2022, Kumarage2023}. These data sources provide a more comprehensive and real-time picture of traffic flows, allowing for more accurate and dynamic O-D matrices. The use of advanced statistical models, machine learning algorithms, and data fusion techniques further enhances the precision of O-D matrix estimations, which not only provide a better understanding of current travel patterns but also enable transportation planners and engineers to predict future trends and make informed decisions for efficient traffic management and sustainable urban planning. Providing O-D matrix optimization would enhance the practicality of the simulation situation.

\begin{figure}[t!]
    \centering
    \includegraphics[width=1\linewidth]{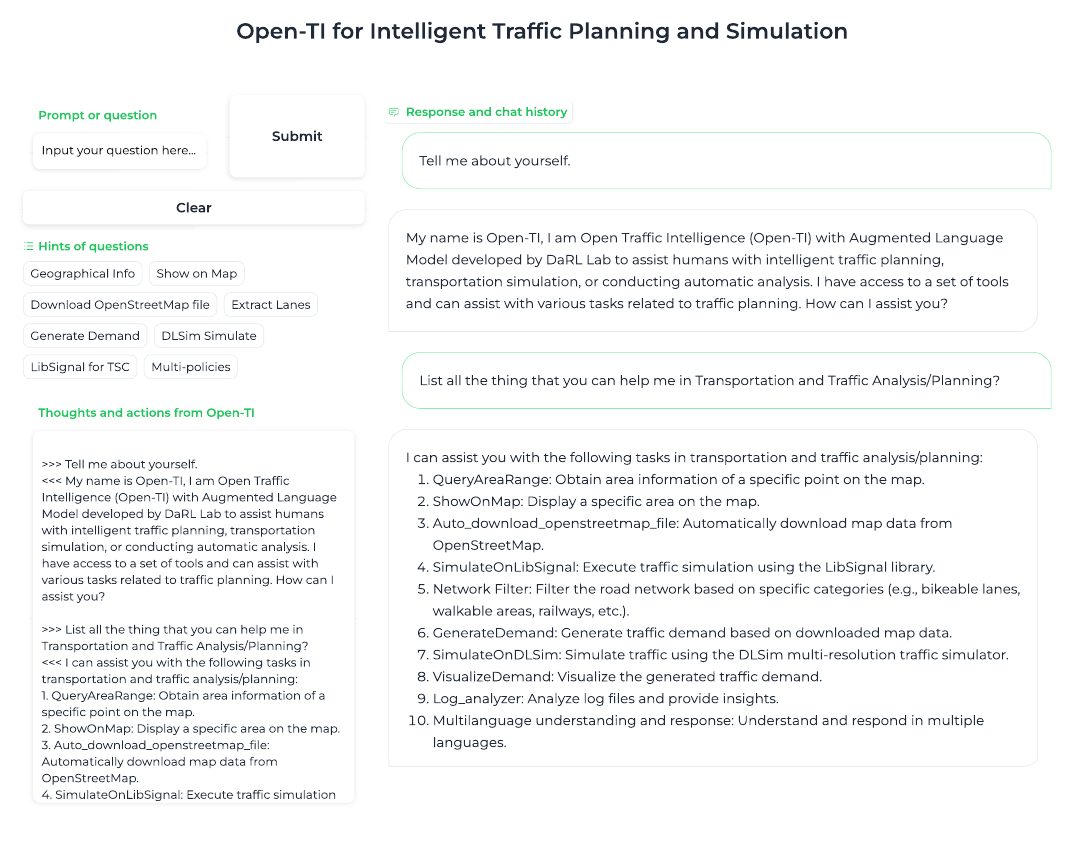}
    \caption{The \ours conversation interface.}
    \label{fig:interface}
\end{figure}

\section{The Architecture of Open-TI}\label{sec3}


\begin{figure}[t!]
    \centering
    \includegraphics[width=0.95\linewidth]{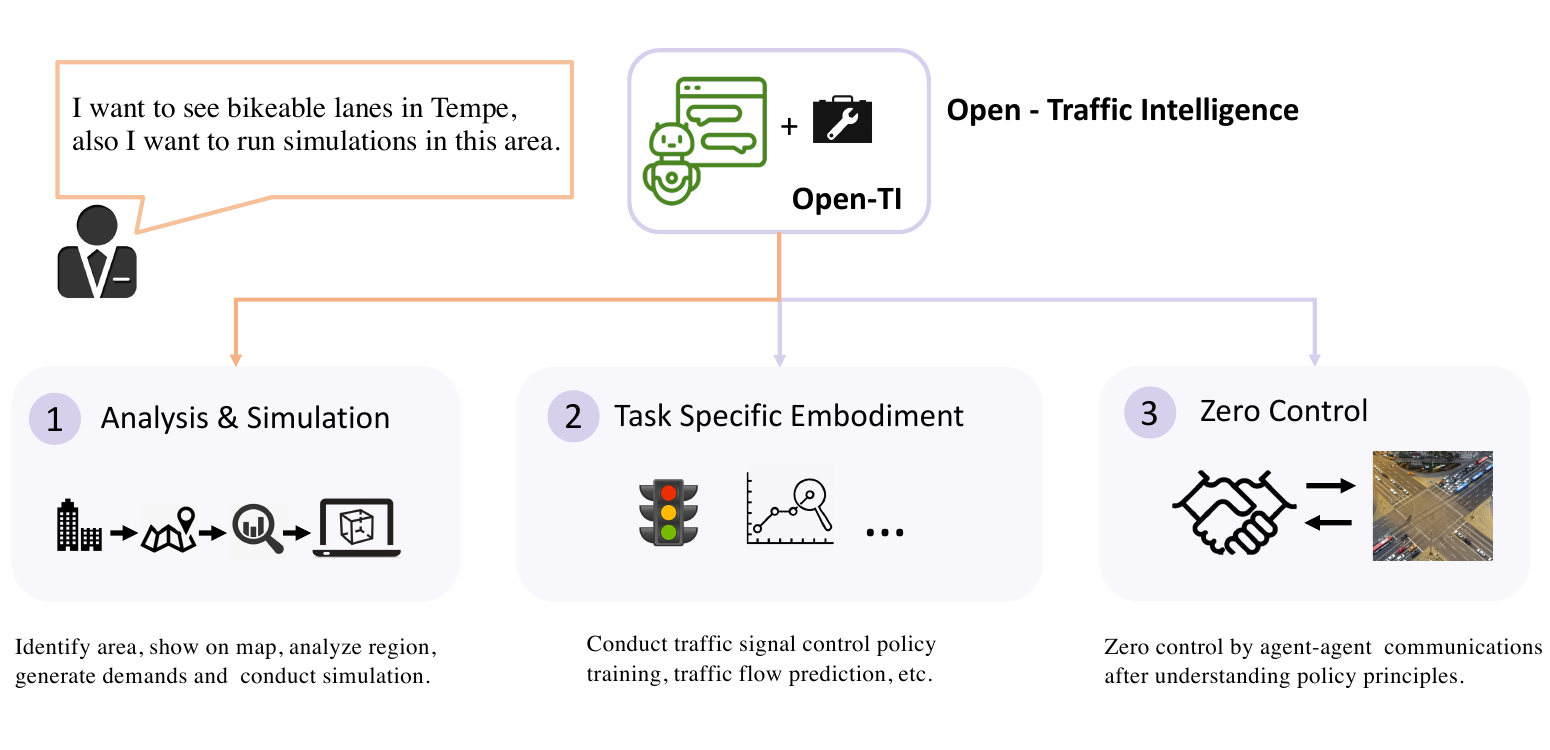}
    \caption{The overview of the \ours functionalities}
    \label{fig:overview}
\end{figure}

\subsection{Overview of \ours}
To take a step forward to more \textbf{T}uring \textbf{I}ndistinguishable \textbf{T}raffic \textbf{I}ntelligence, Open-TI is equipped with human-like semantic ability and goal-oriented execution ability.
Human-like semantic ability is realized by convenient conversation sessions between users and the agent, and execution ability is guaranteed by agent augmentations.

For Open-TI, primarily, a user-friendly interface is designed as shown in Fig.~\ref{fig:interface}. It contains four panels: \textbf{Prompt or question} (top left), the user can edit input, clear text, and submit request; \textbf{Hints of questions} (middle left): user could click on the suggested choices to easily start a conversation; \textbf{Thought and action} (bottom left): this panel presents the chain of thought content from Open-TI agent; \textbf{Response and chat history} (right): this main panel provides multi-media feedback and execution result from Open-TI, including texts, images, path files, and browser links, etc.  

The core of Open-TI mainly incorporates three modules: \textbf{Analysis and Simulation}, \textbf{Task Specific Embodiment} and \textbf{Zero Control} to enhance the intelligent traffic analysis and planning, as shown in Fig.~\ref{fig:overview}. 

First, \ours can manipulate and help practitioners to conduct analysis and simulation from scratch. \ours provides the chance for users to think of a POI (point of interest) or AOI (area of interest) and present the visualization immediately on a map, users can ask for more geology information like latitude and longitude range, and after that, the acquired information can be used to select an analysis range for further investigations like specific lane (e.g., bike lane) filtering and traffic simulation by DTALite (DLSim)~\cite{tong2019open} or SUMO\cite{behrisch2011sumo}. 

Second, the \ours supports multiple task-specific embodiments by vague and high-level language explanations, which greatly reduce the professional background requirements of experimental exploration. For example, based on the current road map, it allows one to conduct traffic light control exploration by either rule-based methods or deep learning policy training ~\cite{mei2023libsignal}, it also could easily conduct traffic demand optimization by providing brief task descriptions.

Third, our method leverages the superior understanding capabilities of current LLMs to conduct meta-control by ChatLight agent: LLMs directly serve as a control agent, follow the understanding of the semantic description of the rules, and control the traffic light actions. This explores a higher level traffic light control mode, e.g., the traffic management department may have special requirements on the safety concerns and would like to adjust the traffic signal control policies to reduce the collision rate, only word-level description is required, and the description would be analyzed, extracted, and communicated as a message to the ChatLight agent, which could reduce the complexity of adjusting the control policy and furthermore, provide explanation for its actions by self-reflecting the LLM's own behavior. 

\begin{figure}[h!]
    \centering
    \includegraphics[width=0.95\linewidth]{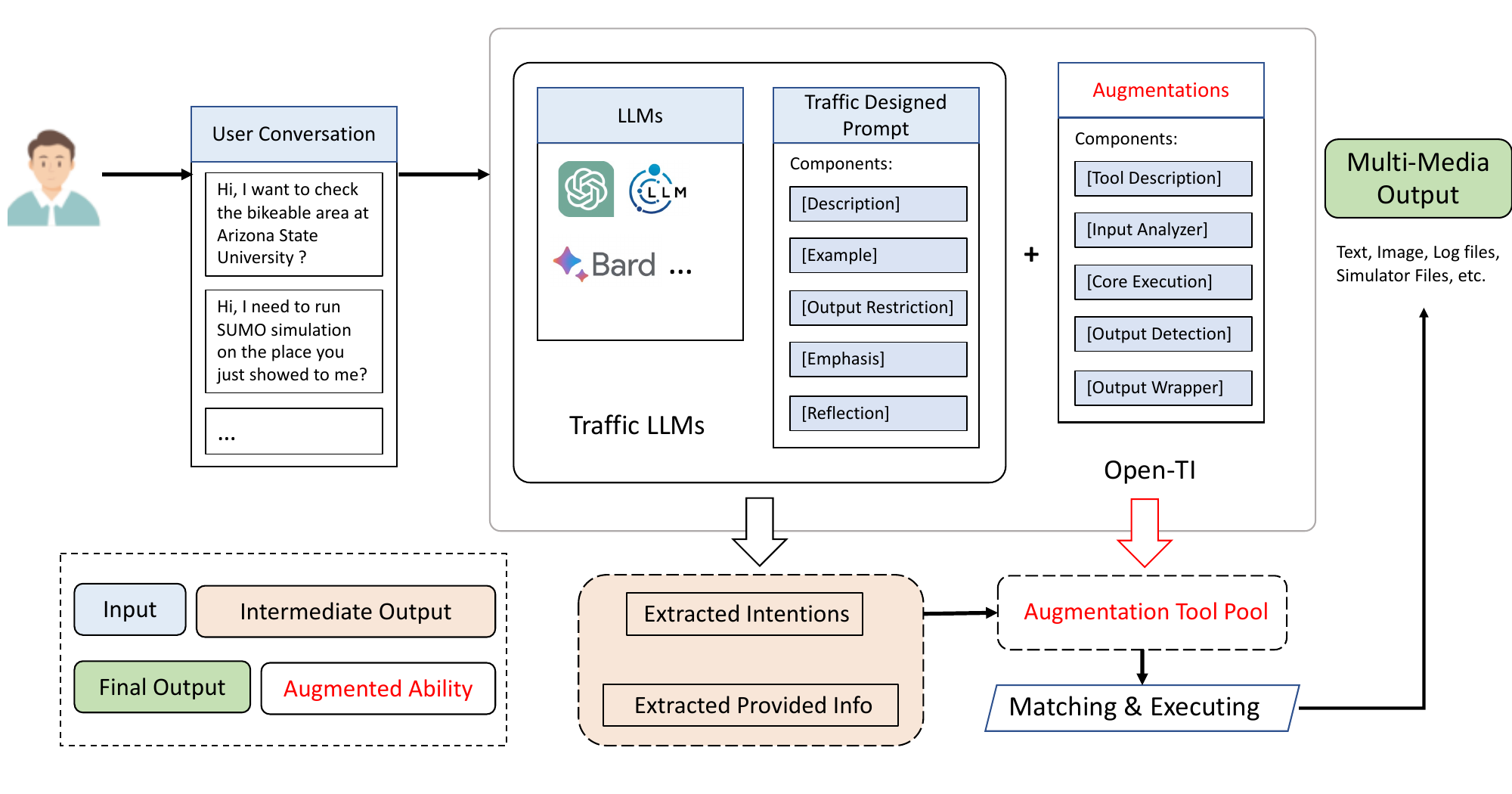}
    \caption{The design framework of \ours}
    \label{fig:framework}
\end{figure}

As shown in Fig.~\ref{fig:framework}, the \ours consists of two components, the Traffic LLMs and Augmentations. When a user requirement is detected from the conversation,  it will be passed to the first components of Traffic LLMs, which use the language agent to extract accurate intentions and useful information from dialogues, then, the extracted intention will be compared with tools in augmentation pool, after matching, if the agent found a possible solution, it will execute with extracted information and generate the output with multimedia data form.

\begin{table}[h!]

\centering
\caption{The details of the prompt components for \texttt{queryAreaRange}}
\label{tab:promptInfo}
\small
\resizebox{0.95\textwidth}{!}{
\begin{tabular}{|c|c|c|}
\hline
\textbf{Name}                                                  & \textbf{Purpose}   &   \textbf{Instances}                                                                                                                                                                                                                                                                                                                                                                                                                                                    \\ \hline
\textbf{\textit{Description}} & \begin{tabular}[c]{@{}c@{}}Core component of the\\ prompt structure, clarifies the basic setup,\\method, and object of each function. \end{tabular}  &  \begin{tabular}[c]{@{}c@{}} You are designed to respond with\\ longitudes and latitudes information of a location\\ Humans might ask similar words of location\\ like position = place = location = geographic info,\\ you can imagine and infer the most possible.     \end{tabular}                                                                                                                                                                                                                                                 \\ \hline
\textbf{\textit{Format Restriction}}             & \begin{tabular}[c]{@{}c@{}}Specify input format constraints,\\ significantly reducing error rates. \end{tabular} &
\begin{tabular}[c]{@{}c@{}} The format of your output longitude and \\latitude is a query of 4 value array as\\ {[}min\_long, min\_lat,max\_long,max\_lat{]}\end{tabular}
\\ \hline
\textbf{\textit{Example}}        & \begin{tabular}[c]{@{}c@{}}Help LLMs understand the\\ exactly processing of the execution. \end{tabular} &\begin{tabular}[c]{@{}c@{}}Human ask ``Where is Arizona State University,\\ Tempe Campus", you need to output\\ {[}-111.9431, 33.4154, -111.9239, 33.4280{]}.\end{tabular}\\ \hline

\textbf{\textit{Reflection}}           & \begin{tabular}[c]{@{}c@{}}Remind LLMs not to engage in \\ unnecessary tasks and ensure \\that each process is executed accurately. \end{tabular}  &  \begin{tabular}[c]{@{}c@{}}You should respond directly with what\\ you know and then stop, do not look for \\the location or attempt to find it online. \end{tabular}                                                                                                                                                                                                                                                       \\ \hline
\textbf{\textit{Emphasis}}            & \begin{tabular}[c]{@{}c@{}}Reinforce the function's objective, \\significantly reducing API mismatching rates.\end{tabular} &\begin{tabular}[c]{@{}c@{}}You have the valid tool to provide location. You \\have a specific tool to directly query the location.\end{tabular}                                                                                                                                                                                                                                                                                              \\ \hline
\end{tabular}
}
\end{table}

\subsection{Prompt Design}
The \ours exploited the LLM's understanding abilities and tuned its behavior by prompt engineering, which is essential for well-organized operation. We have designed 5 aspects of prompt structure:~\textbf{[\textit{Description}]}, \textbf{[\textit{Example}]}, \textbf{[\textit{Output Restriction}]}, \textbf{[\textit{Emphasis}]} and \textbf{[\textit{Reflection}]}. And we verified their effectiveness in cross-task behaviors by ablation experiment. In this section, we provide details of prompt design as shown in Fig.~\ref{fig:framework}. Including the purpose of the prompt and example cases. The examples in Table~\ref{tab:promptInfo}. are from the same augmentation task of \texttt{queryAreaRange}.

\subsection{Execution and Augmentation List}
The overall execution process is expressed in Algorithm~\ref{alg:process}. in pseudo-code. Following the same execution flow, there are different augmented tools that could help users with various requirements as presented in Table~\ref{tab: augment_info}. The section.~\ref{sec:embodiment} will elaborate on the three augmentation modules in detail.
\begin{table}[h]
\centering
\caption{A list of augmented tools implemented in \ours}
\label{tab: augment_info}
\resizebox{0.9\textwidth}{!}{
\begin{tabular}{|c|c|}
\hline
\textbf{Augmentation Name} & \textbf{Description}      \\ 
\hline
\texttt{\texttt{queryAreaRange}}      & \begin{tabular}[c]{@{}c@{}}Obtain area information, specifically the\\ longitudes and latitudes of a point of interest on the map.\end{tabular}    \\ 
\hline
\texttt{\texttt{showOnMap}}      & \begin{tabular}[c]{@{}c@{}}Display the location of interest \\on the map, such as the ASU campus area.  \end{tabular}  \\ 
\hline
\texttt{\texttt{autoDownloadOpenStreetMapFile}}   &\begin{tabular}[c]{@{}c@{}} Automatically download map data \\from OpenStreetMap for a specified area. \end{tabular}\\
\hline
\texttt{\texttt{simulateOnLibsignal}}       &   \begin{tabular}[c]{@{}c@{}}Execute simulations on the\\ open-source library called Libsignal.\end{tabular} \\ 
\hline
\texttt{\texttt{networkFilter}} & \begin{tabular}[c]{@{}c@{}}Filter the road network based on\\ required categories, return the file\\ path of a filtered road network \\that emphasizes lanes of interest.\end{tabular} \\ 
\hline
\texttt{\texttt{generateDemand}}         & \begin{tabular}[c]{@{}c@{}} Generate demand based on \\OpenStreetMap data. \end{tabular}\\ 
\hline
\texttt{\texttt{simulateOnDLSim}}          &  \begin{tabular}[c]{@{}c@{}}Simulate on the DLSim\\ multi-resolution traffic simulator. \end{tabular}  \\ 
\hline
\texttt{\texttt{simulateOnSUMO}}          &  \begin{tabular}[c]{@{}c@{}}Execute the simulation given \\arbitrary .osm data. \end{tabular} \\ 
\hline
\texttt{\texttt{visualizeDemand}}          &   \begin{tabular}[c]{@{}c@{}}Automatically generate and display\\visualizations of the demand file.\end{tabular}\\ 
\hline
\texttt{\texttt{logAnalyzer}}              & \begin{tabular}[c]{@{}c@{}}Analyze log or config files\\ and provide comparisons. \end{tabular}   \\
\hline
\texttt{\texttt{resultExplainer}}              &\begin{tabular}[c]{@{}c@{}} Interpreter results to provide insights.  \end{tabular}  \\
\hline
\texttt{\texttt{demandOptimizer}}              & \begin{tabular}[c]{@{}c@{}}Approximate the origin-destination \\demand to fit realistic observation. \end{tabular}  \\

\hline
\end{tabular}
}
\end{table}

\begin{algorithm}
\caption{Open-TI Execution Process}
\label{alg:process}
\begin{algorithmic}[1] 
\relscale{0.7}
\State INPUT: $msg \gets \textit{UserInputQuery}$
\If{Intention Needs External Tools}
    \While{Augmented APIs not Found}
        \State $keywords \gets \text{summarize}(msg)$
        \State $api \gets \text{search}(keywords)$
        \If{MaximumQueryTime Exceeds}
            \State \textbf{break}
        \EndIf
    \EndWhile
    \If{API found}
        \State \textit{Params} $\gets$ extract\_params(\textit{msg})
        \While{\textit{Params} Not Satisfied}
            \State 1. Retrospect Expected Form
            \State 2. Examine User Input: \textit{msg}
            \While{Missing Info.}
                \State Alert Required Info.
            \EndWhile
            \State $Response \gets \text{execute\_api\_call}(\textit{Params})$
            \If{MaximumQueryTime Exceeds}
                \State \textbf{break}
            \EndIf
        \EndWhile
        \While{Response not Satisfied}
            \State $api\_call \gets \text{gen\_api\_call}(api\_doc, msg)$
            \State $Response \gets \text{execute\_api\_call}(api\_call)$
            \If{MaximumQueryTime Exceeds}
                \State \textbf{break}
            \EndIf
        \EndWhile
    \EndIf
\EndIf
\If{Response}
    \State $re \gets \text{Construct\_Response}(Response)$
\Else
    \State $re \gets \text{Query\_Failed}()$
\EndIf
\State \Return $re \gets \text{ResponseToUser}$
\end{algorithmic}
\end{algorithm}

\subsection{Standard Implementation}
In this section, we formalize the API content and provide the structure of the augmentation class. In order to extend the augmentation to the border range with additional capabilities, the keys are: First, make sure the pivotal agent is aware of the functionality of the specific additional tool, so when the requirement is asked by users, the pivotal agent could pick up the most suitable method. Second, the embodiment program should function accurately and effectively, so that when the pivotal agent boots up the process, it can come to the expected result. Based on the two keys, we provide the implementation structure as in the following code. One needs to provide the descriptions on the augmentation class and implement the \texttt{Execution ()}, this provides an easy way to understand and neat for other researchers to explore and enforce Open-TI's abilities.
\begin{center}
\begin{minipage}{.95\textwidth}
\label{lis:code}
\begin{lstlisting}[language=Python, caption=The Class and API Format , label=code:api-example]
class AugmentTask:
    def __init__(self, params) -> None:
        self.params = params
        # set possible pre-defined params, e.g., folder path

    @func_prompt(name="prompt name",
        description="""detailed explanation""")

    def embody(self, target: str) -> str:
        try:
            result = Execution(target)
            # Concrete Execution Implementation
        except Error as e:
        print(e)
        return result
            
\end{lstlisting}
\end{minipage}
\end{center}

\section{Sub-module Embodiment}\label{sec:embodiment}
The \ours is enhanced by three different modules that eventually realize the superior performance. In this section, we will elaborate on each module with examples and illustrations. 

\subsection{Pivotal Agent for Transportation Analysis}\label{subsubsec2}

\begin{figure}[h!]
    \centering
    \includegraphics[width=0.99\linewidth]{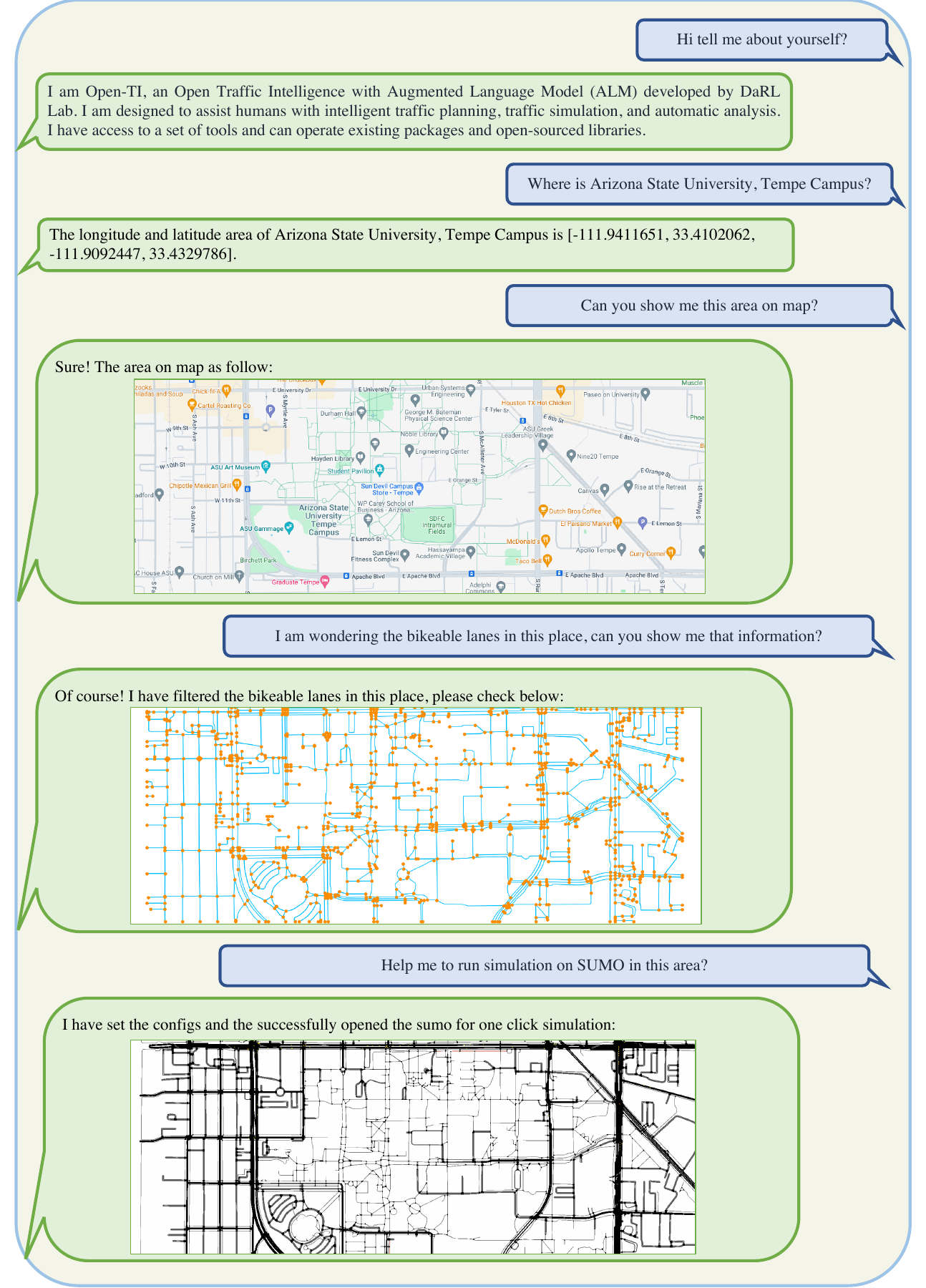}
    \caption{The demonstration pivotal agent control. (Right: The user messages, Left: The responses from \ours). This series of interactions shows how to query geography information of a location, how to visualize on the map, filter the interested lane types, and use the arbitrary map for automatic traffic simulation (SUMO).}
    \label{fig:interactdemon}
\end{figure}

In this module, analysis from scratch is realized by seamless connections between augmented tools and the pivotal operation agent. The supported external tools and packages are shown in Table~\ref{tab:tools}. And when the user asks about related tasks, \ours will automatically match the highest probability option and process following Algorithm 1. An example of interaction is shown in Fig.~\ref{fig:interactdemon}.

\begin{table}[htb]
\small
\centering
\caption{The supported external tools and packages}
\label{tab:tools}
\setlength{\tabcolsep}{1mm}
    \begin{tabular}{cccccc}
    \toprule
    Name & Functions & Versions      \\ \midrule
    osm2gmns & obtain networks from OSM and convert to GMNS & V-0.7.3  \\
    grid2demand   &   Origin-destination trans demand generate     &  V-0.3.6 \\
    DLSim-MRM   &   Multi-resolution Traffic Simulation     &  V-0.2.11  \\
    Libsignal   &   Multi-simulator platform for Traffic Signal Control     &  V-1.0.0 \\
    
    \bottomrule
    \end{tabular}
\end{table}

\subsection{Task-Specific Embodiment }\label{subsubsec2}
The \ours is capable of realizing more general research tasks in the traffic domain. Including traffic signal control (TSC), traffic flow prediction, traffic Origin-Destination(O-D) demand optimization, etc. The architecture in \ours is well structured and supports extensibility for an open community. We will introduce how the \ours achieves the three demonstrating tasks in the following subsections.


\subsubsection{Traffic O-D Demand Optimization Task}

\begin{figure}[h!]
    \centering
    \begin{tabular}{cc}
    \includegraphics[width=0.450\textwidth]    
    {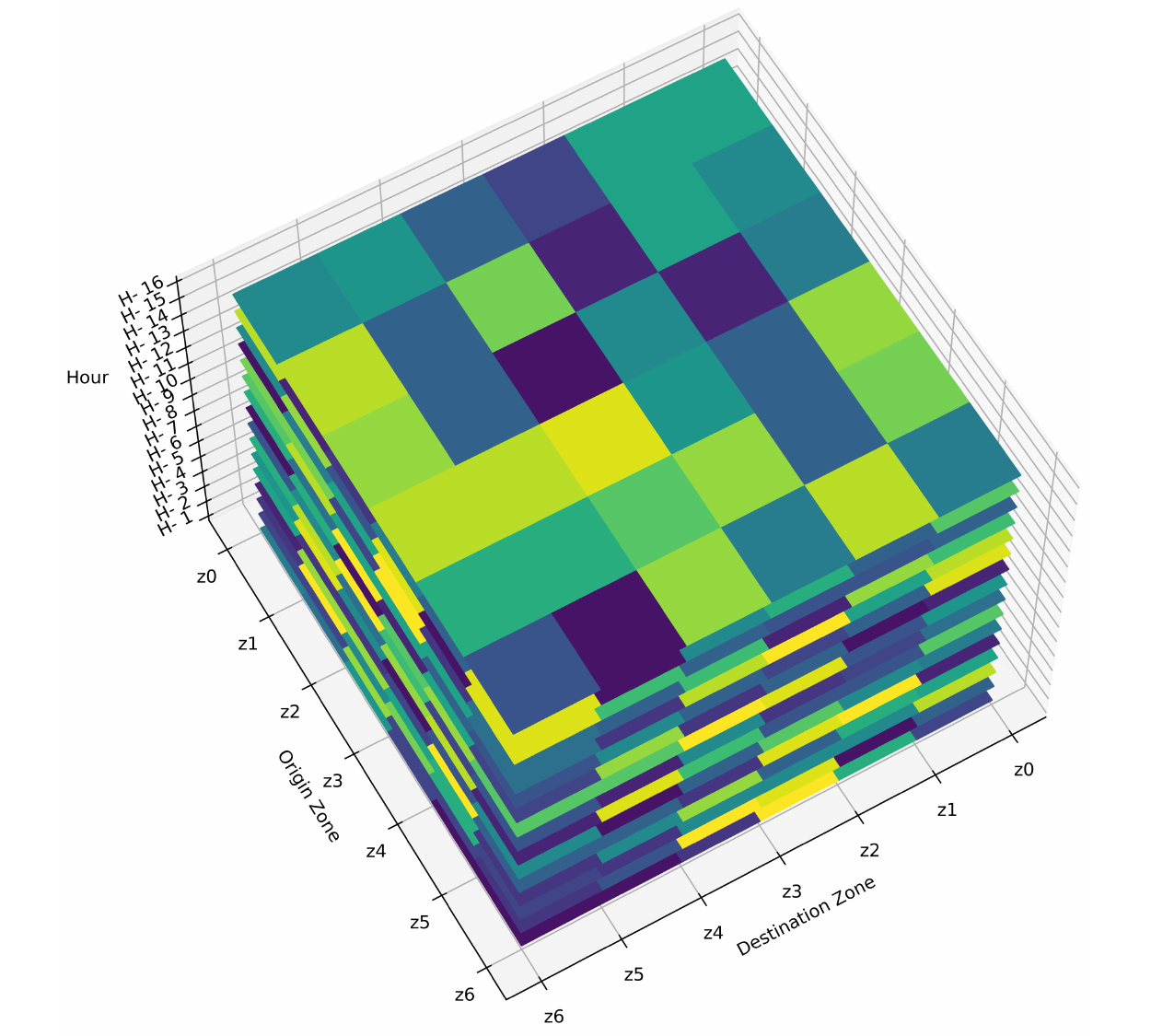}
    &\includegraphics[width=0.460\textwidth]
    {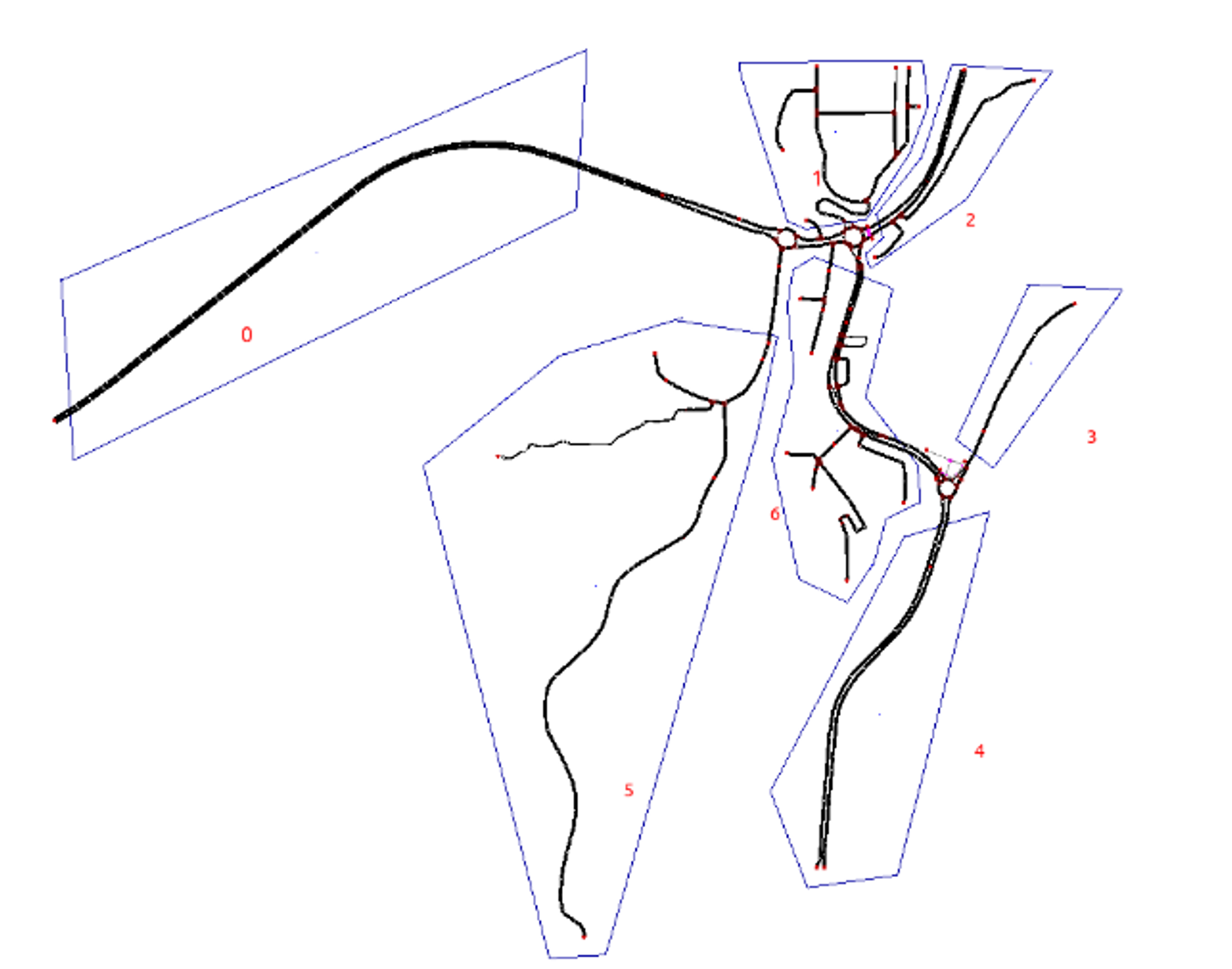} \\
       (a) The visualization of O-D matrix & 
       (b) The demand zones split
    \end{tabular}
    \caption{An illustration of the O-D demand optimization task in Sedona, AZ. In (a), it consists of 3 dimensions: origin zone, destination zone, and time (Hour), in each time interval, one slice describes the traffic demand information. (b) is the geography demand zone split of the interested area.}
    \label{fig:od-demand}
\end{figure}
In the OD demand optimization task, the goal is to design an algorithm and learn a model that could help to output an accurate OD matrix, given the partial observation data. In Figure~\ref{fig:od-demand}, we show an example in Sedona, AZ, USA. When asked to execute an O-D matrix optimization task, users could specify the observation data source, and traffic simulation setting, and then choose the optimization techniques to experiment with. In the example case, the given data is the 16-hour count data at the specific observation point of a roundabout, and we asked the agent to use a genetic algorithm to conduct optimization and provide the result.


\subsubsection{Traffic Signal Control Task}
In the realization of traffic signal control embodiment, we seamlessly integrated the Libsignal~\cite{mei2023libsignal} that could realize the cross-simulator traffic signal control over the majority of baseline methods, including the rule-based approaches (Fixed Time and Self-organizing traffic lights - SOTL~\cite{cools2013self}) and reinforcement-learning-based approaches as shown in Fig.~\ref{fig:tscmatch}. We provide further interaction cases in the Appendix.~\ref{app:example1}.
\begin{figure}[h!]
    \centering
    \includegraphics[width=0.95\linewidth]{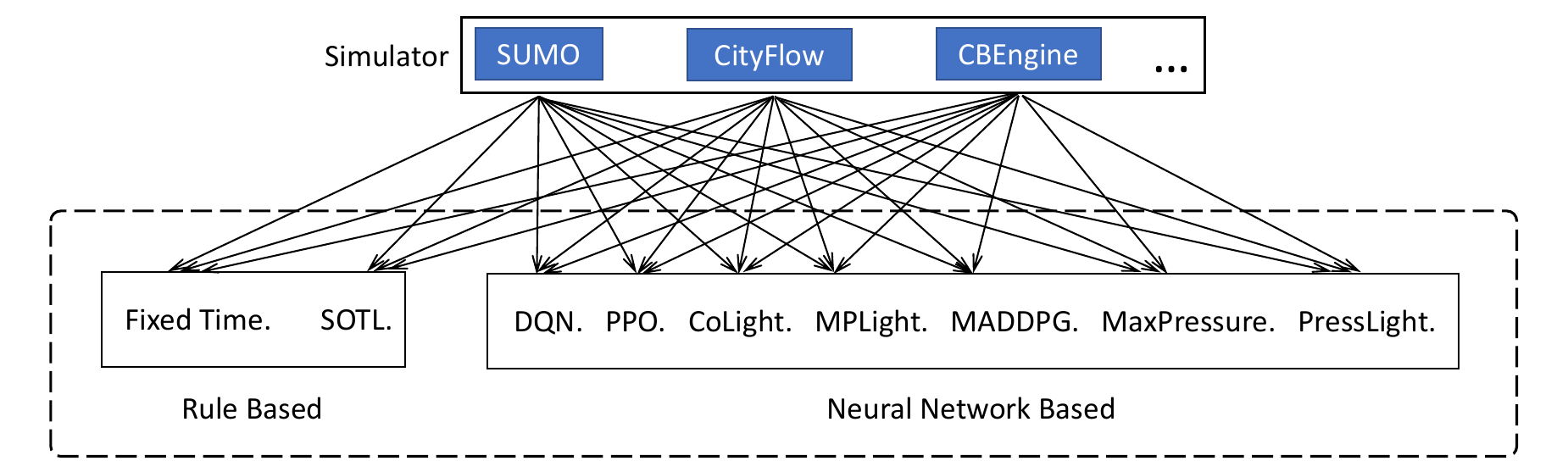}
    \caption{Current feasible simulators and algorithms for TSC tasks}
    \label{fig:tscmatch}
\end{figure}

\subsection{Agent Meta Control}\label{subsubsec2}
In the meta-control task, we essentially designed an agent-agent communication and the pivotal agent is in charge of understanding the human descriptive intention of the traffic policy, and the execution agent will take the message as instruction, process, and digest, then connect to the traffic practical control to provide a self-explainable action execution control.
\begin{figure}[h!]
    \centering
    \includegraphics[width=0.95\linewidth]{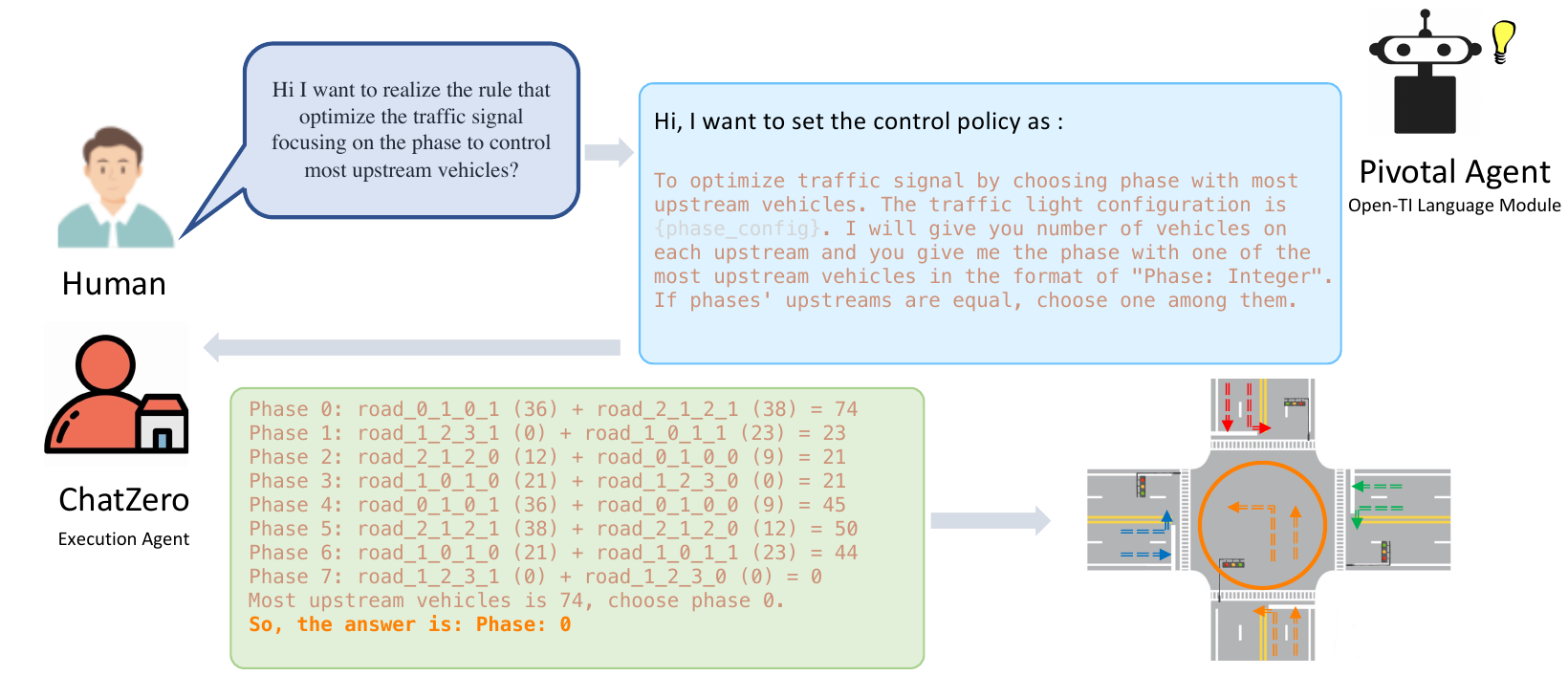}
    \caption{The demonstration ChatZero Meta Control.}
    \label{fig:metacontrol}
\end{figure}

\section{Experiment}\label{sec4}
In this section, we conduct extensive experiments to answer the following questions:
\begin{itemize}
    \item RQ1: How does our proposed approach compare to other state-of-the-art methods in terms of performance?
    \item RQ2: How do the components impact the performance of our proposed method?
    \item RQ3: How does ChatZero execute meta-control? How does it perform across various LLMs?
\end{itemize}

Three aspects of experiments are designed from error rates of API calls, ablation study of Augmented Languages Agents' prompt structure, and Zero-Control agent's performance on LLM agents to verify the effectiveness and stability of the proposed \ours. Please note that, for the RQ1 and RQ2, we develop standard \ours based on GPT3.5 and for RQ3, we verified on 4 different language models: Llama7B, Llama13B, GPT3.5, and GPT4.0.

\subsection{Language Agent Analysis on the API Calls}

In this section, we conduct the functionality-level experiments of API analysis and compare them with the baseline method known as TrafficGPT\cite{zhang2023trafficgpt}.

\paragraph{Experiment Design} First, following the work of~\cite{li2023api}, we analyze three types of API call abnormal behaviors, namely `No API Call Rate', `API Mismatching Rate', and `Error Raise Rate'.  Both \ours and TrafficGPT are equipped to handle a range of tasks spanning geographical information, simulation, and traffic signal control. Although the specific functions of \ours and TrafficGPT are slightly different, we are still able to evaluate the overall API access stability. We adopted $T = 6$ similar tasks as shown in the table~\ref{tab: compare_task}. to conduct the comparison by testing each task 20 times. And calculate the error rate follow the equation~\ref{eq:error}:

\begin{equation}
    \textit{Error Rate} = \frac{1}{T}\sum_{t=1}^{T} \frac{n_{t}^e}{n_{t}^c}=\frac{1}{T}\sum_{t=1}^{T} \frac{1}{n_{t}^c}\sum(c_t^{no},c_t^{miss},c_t^{error})
    \label{eq:error}
\end{equation}

where $n_t^{e}$ represents the number of error occurrences for task $t$ during total tests, $n_{c}$ denotes the number of total testing instances, (i.e., $n_{c} = 20 $ for this experiment), $c_t^{no}$ is the sum of errors caused by the absence of API calls for task $t$ among all tests, similarly, $c_t^{miss}$ is the sum of mismatching error times, $c_t^{error}$ is the number of error raising times, and exist $n_t^{e} = c_t^{no} + c_t^{miss} + c_t^{error}$.

\begin{figure}[h!]
    \centering
    \includegraphics[width=0.6\linewidth]{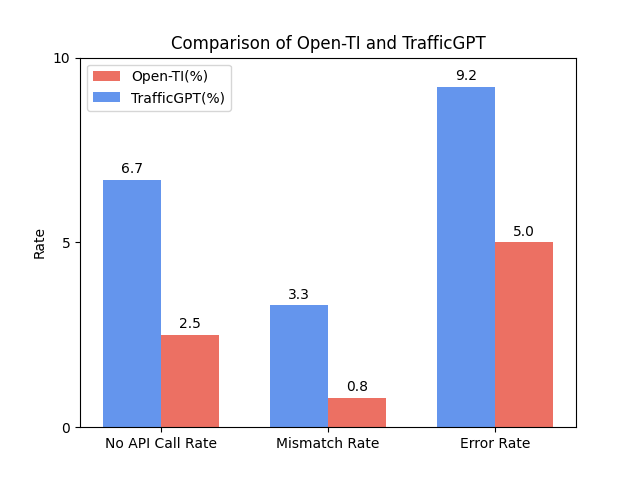}
    \caption{Language agent analysis on the API calls}
    \label{fig:apicall}
\end{figure}

For evaluation of each dimension shown in Fig.~\ref{fig:apicall}, denote $\rho$ as error rate, we have: 
$\rho_{no} =  \frac{1}{T}\sum_{t=1}^{T} \frac{c_t^{no}}{n_{t}^c}$, $\rho_{miss} =  \frac{1}{T}\sum_{t=1}^{T} \frac{c_t^{miss}}{n_{t}^c}$ and $\rho_{error} =  \frac{1}{T}\sum_{t=1}^{T} \frac{c_t^{error}}{n_{t}^c}$.

\begin{table}[h]
\small
\centering
\caption{The compared tasks is an intersection set that exists both in \ours and TrafficGPT. The experiment tends to design a fair comparison by identical task goals or similar difficulty.}
\label{tab: compare_task}
\begin{tabular}{|c|c|c|}
\hline
 & \textbf{\ours} & \textbf{TrafficGPT}      \\ 
\hline
\ding{172}&\texttt{simulateOnSumo}      & Run the sumo simulation\\ 
\hline
\ding{173}&\texttt{showOnMap}      & Draw intersections on map    \\ 
\hline
\ding{174}&\texttt{logAnalyzer}    & Retrieve data from the .xml files \\
\hline
\ding{175}&\texttt{visualizeTrainingCurves} & Generating heat graphs \\ 
\hline
\ding{176}&\texttt{simulateOnLibsignal}  &  Optimize intersections by Webster \\ 
\hline
\ding{177}&\texttt{resultExplainer}      &  Compare data from the simulation \\ 
\hline
&\texttt{queryRangeArea}      &  - \\ 
\hline
&\texttt{autoDownloadOpenStreetMapFile}      &  - \\ 
\hline
&\texttt{networkFilter}      &  - \\ 
\hline
&\texttt{generateDemand}      &  - \\ 
\hline
&\texttt{simulateOnDLSim}      &  - \\ 
\hline
&\texttt{demandOptimizer}      &  - \\ 
\hline

\end{tabular}
\end{table}

As in the top of Table \ref{tab: compare_task}, which is an intersection set of functionalities in TrafficGPT and \ours. \texttt{simulateOnSumo} are commonly integrated by different implementations, \texttt{showOnMap} is used to query and show the map of the interested place which is equivalent to the functionality of locating and drawing the intersection on the map. \texttt{logAnalyzer} is designed to interpret and help the user understand the log and config files, similar to retrieving data from .XML files in TrafficGPT. \texttt{visualizeTrainingCurves} is for visualization of the training process which is equivalent to generating heat graphs by TrafficGPT, \texttt{simulateOnLibsignal} applies multiple algorithms to control traffic signal control while the Webster method is used in TrafficGPT, similarly, two systems both provide result explanation as \texttt{resultExplainer}, which we take into comparison.

\paragraph{Experimental Results}
\begin{table}[h!]
\centering
\caption{Error Rate of \ours and TrafficGPT}
\resizebox{0.5\textwidth}{!}{
\begin{tabular}{|cc|c|c|}
\hline
\multicolumn{2}{|c|}{\textbf{Tasks}}                             & \textbf{TrafficGPT} & \textbf{Open-TI} \\ \hline
\multicolumn{1}{|c|}{\multirow{3}{*}{\ding{172}}} & No API Call & \cellcolor[HTML]{cae6f2}0.00  & \cellcolor[HTML]{cae6f2}0.00             \\ \cline{2-4} 
\multicolumn{1}{|c|}{}& Mismatch & \cellcolor[HTML]{6fb9dc}0.05                 & \cellcolor[HTML]{cae6f2}0.00                \\ \cline{2-4} 
\multicolumn{1}{|c|}{}                    & Error Raise & \cellcolor[HTML]{2190c5}0.10  & \cellcolor[HTML]{6fb9dc}0.05 \\ 
\hline
\multicolumn{1}{|c|}{\multirow{3}{*}{\ding{173}}} & No API Call & \cellcolor[HTML]{6fb9dc}0.05                 & \cellcolor[HTML]{cae6f2}0.00                \\ \cline{2-4} 
\multicolumn{1}{|c|}{}                    & Mismatch    & \cellcolor[HTML]{cae6f2}0.00                    & \cellcolor[HTML]{cae6f2}0.00                \\ \cline{2-4} 
\multicolumn{1}{|c|}{}                    & Error Raise & \cellcolor[HTML]{cae6f2}0.00                    & \cellcolor[HTML]{6fb9dc}0.05             \\ \hline
\multicolumn{1}{|c|}{\multirow{3}{*}{\ding{174}}} & No API Call & \cellcolor[HTML]{006696}0.15                 & \cellcolor[HTML]{cae6f2}0.00                \\ \cline{2-4} 
\multicolumn{1}{|c|}{}                    & Mismatch    & \cellcolor[HTML]{6fb9dc}0.05                 & \cellcolor[HTML]{6fb9dc}0.05             \\ \cline{2-4} 
\multicolumn{1}{|c|}{}                    & Error Raise & \cellcolor[HTML]{2190c5}0.10                  & \cellcolor[HTML]{6fb9dc}0.05             \\ \hline
\multicolumn{1}{|c|}{\multirow{3}{*}{\ding{175}}} & No API Call & \cellcolor[HTML]{6fb9dc}0.05                 & \cellcolor[HTML]{cae6f2}0.00                \\ \cline{2-4} 
\multicolumn{1}{|c|}{}                    & Mismatch    & \cellcolor[HTML]{2190c5}0.10                  & \cellcolor[HTML]{cae6f2}0.00                \\ \cline{2-4} 
\multicolumn{1}{|c|}{}                    & Error Raise & \cellcolor[HTML]{006696}0.15                 & \cellcolor[HTML]{2190c5}0.10              \\ \hline
\multicolumn{1}{|c|}{\multirow{3}{*}{\ding{176}}} & No API Call & \cellcolor[HTML]{6fb9dc}0.05                 & \cellcolor[HTML]{6fb9dc}0.05             \\ \cline{2-4} 
\multicolumn{1}{|c|}{}                    & Mismatch    & \cellcolor[HTML]{cae6f2}0.00                    & \cellcolor[HTML]{cae6f2}0.00                \\ \cline{2-4} 
\multicolumn{1}{|c|}{}                    & Error Raise & \cellcolor[HTML]{2190c5}0.10                  & \cellcolor[HTML]{6fb9dc}0.05             \\ \hline
\multicolumn{1}{|c|}{\multirow{3}{*}{\ding{177}}} & No API Call & \cellcolor[HTML]{2190c5}0.10                  & \cellcolor[HTML]{6fb9dc}0.05             \\ \cline{2-4} 
\multicolumn{1}{|c|}{}                    & Mismatch    & \cellcolor[HTML]{cae6f2}0.00                    & \cellcolor[HTML]{cae6f2}0.00                \\ \cline{2-4} 
\multicolumn{1}{|c|}{}                    & Error Raise & \cellcolor[HTML]{2190c5}0.10                  & \cellcolor[HTML]{cae6f2}0.00                \\ \hline
\end{tabular}
}\label{tab:sidebyside}
\end{table}

The experiment results can be found in Fig.~\ref{fig:apicall}. The comparison is conducted on the average value over 20 times. The sum of the 3 types of error rates in \ours and TrafficGPT are 8.3\% and 19.2\%, calculated by aggregation of the three types of error rates in two systems respectively.

In Fig.~\ref{fig:apicall}, we take the $\textit{mean}$ as the evidence to show on the bar chart as from 3 different evaluation dimensions. The x-axis presents individual dimensions of error rates. In each dimension, the two methods are compared side by side, and we can compare their percentage value according to the y-axis. 

We could observe that the \ours effectively reduced the general API abnormal behaviors across three evaluation aspects. More obviously, \ours shows significantly better performance than the baseline method in terms of Error Rate, mainly because the 5 aspects of the prompt components emphasize the task, reactions, and meta-information more explicitly. We also demonstrate the detailed task level side-by-side comparison as shown in Table.~\ref{tab:sidebyside}, from the result, we could notice that \ours performs consistently more stable (lower error rate and lighter color) for each task than TrafficGPT, even though for task \ding{173}, Open-TI performs slightly worse in Error Raise Rate, it is because \texttt{showOnMap} is a more complex task consists of two consecutive sub-tasks, which are 1) identifying the geographic information and 2) request and visualize map data.  To understand how each component plays a role in helping the language agent work consistently stable, we conduct further exploration in the next section.


\subsection{Ablation Study of \ours Prompt Structure}

In this section, we conduct the ablation study to unravel the nuanced impact of different components on the functionality of the tasks under consideration.

 

\paragraph{Experiment Design}
The ablation process is operated by the removal of each prompt component in an order of \textbf{\textit{Emphasis}}, \textbf{\textit{Reflection}}, \textbf{\textit{Format Restriction}}, \textbf{\textit{Example}}, and \textbf{\textit{Description}}. The analysis is done across 4 tasks: \texttt{queryAreaRange}, \texttt{showOnMap}, \texttt{autoDownloadOpenStreetMapFile}, and \texttt{simulateOnLibsignal}, which encompasses all five prompt components.

\begin{figure}[h!]
    \centering
    \includegraphics[width=0.7\linewidth]{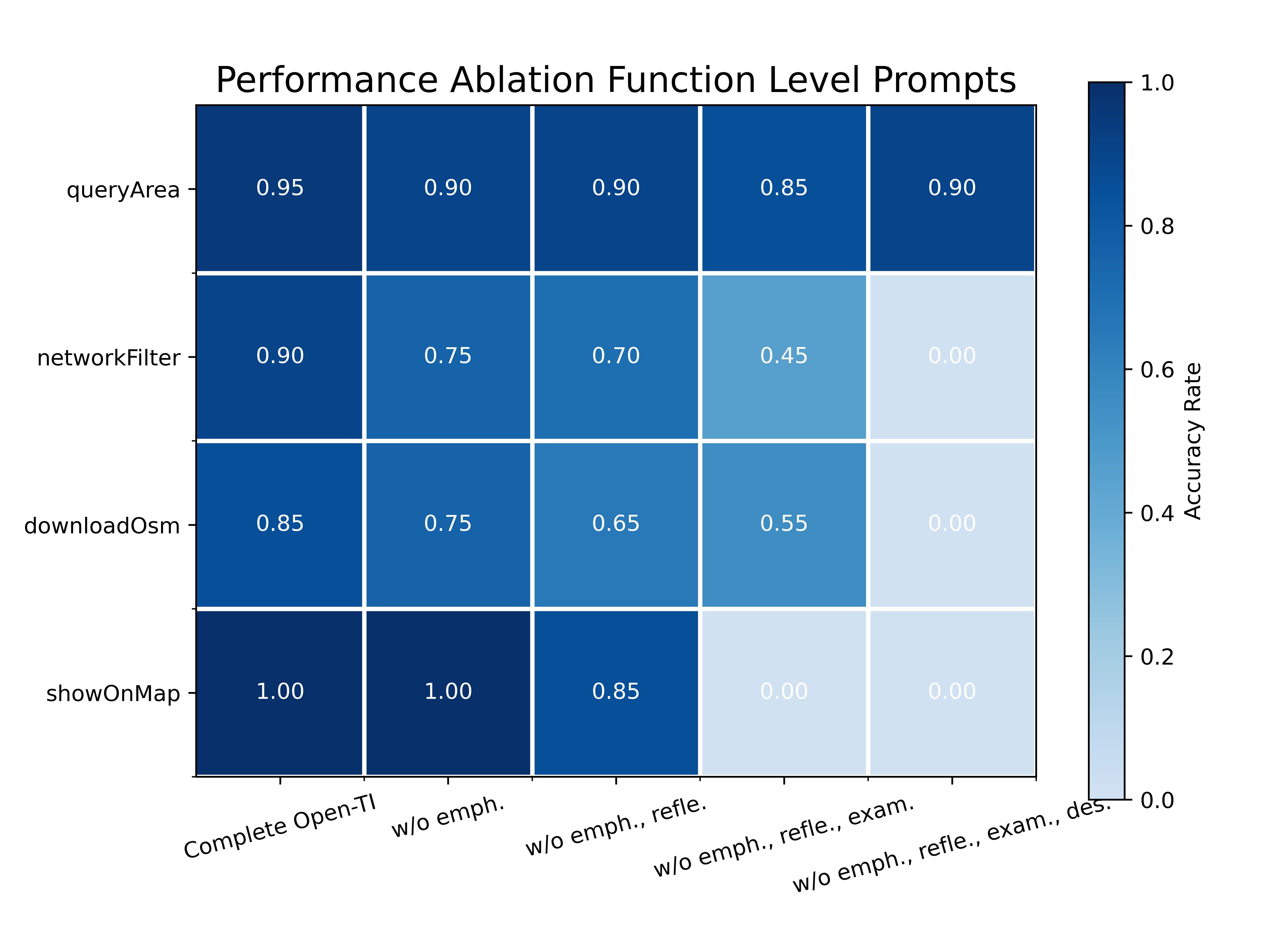}
    \caption{The ablation study of the prompt components. The x-axis from left to right shows the gradual removal of each prompt module. The y-axis shows the individual task, cell color from dark to light reflects the performance drop.} 
    \label{fig:heatmap}
\end{figure}

By eliminating the least important component first, followed by the removal of subsequent components in ascending order of importance. This sequential approach enables the examination of the individual influence of each component on the overall function. For each iteration, we conduct an experiment by posing 20 questions related to each function. Then the collected error rates are summarized into a comprehensive heatmap as in Fig.~\ref{fig:heatmap}.

\paragraph{Experimental Results}
Fig.~\ref{fig:heatmap} shows the experimental findings related to the effect of each component. When there is no component removed, the system's performance is relatively stable and robust.  When the \textbf{\textit{Emphasis}} is removed, the accuracy suffers a slight drop across \texttt{queryAreaRange} to \texttt{autoDownloadOpenStreetMapFile}, because the agent can be confused among similar tasks or tasks with similar keywords then leading to mismatching. For example, when a human asked, "Can you help me find the map of Arizona State University?" The agent may be confused about finding the map and suppose human need the geographical information about Arizona State University. Thus, the agent mismatch \texttt{autoDownloadOpenStreetMapFile} question to \texttt{queryAreaRange}. 

Then if the \textbf{\textit{Reflection}} is further removed, \texttt{simulateOnLibsignal} suffers the most, by investigation, we found that it is mainly caused by entering incorrect keywords in questions. However, the agent can memorize the answers to previous questions, preventing error escalation and subsequently providing inaccurate responses to users. Therefore, the \textbf{\textit{Reflection}} aims to address the problem of incorrect keywords, ensuring that users input the correct keywords in their questions. 

Meanwhile, \textbf{\textit{Format Restriction}} mainly affects \texttt{showOnMap} and \texttt{simulateOnLibsignal}. When \textbf{\textit{Format Restriction}} is removed, their accuracy rates decrease by 25\% and 55\%, respectively. This is due to the role of \textbf{\textit{Format Restriction}} in input format limit control; Entering incorrect information into a project can lead to errors. As a result, \textbf{\textit{Format Restriction}} significantly affects \texttt{showOnMap} and \texttt{autoDownloadOpenStreetMapFile} performance.

The component \textbf{\textit{Example}} plays a significant role in helping language agents understand the task, once removed, the accuracy rates decrease by 45\% and 55\%, respectively. Furthermore, since LLMs can sometimes overlook the middle steps in implementation, the  \textbf{\textit{Example}} helps to improve its stability. Thus, the Example is significant in both \texttt{showOnMap} and \texttt{autoDownloadOpenStreetMapFile}. For the case of overlook behavior, e.g., if I want to download the OSM file of Arizona State University and the agent doesn't have the geographical information for the target place, the agent should first call the \texttt{queryAreaRange} function. After obtaining the geographical information, it should then input this data into the \texttt{autoDownloadOpenStreetMapFile} function to get the correct answer. We define this embodiment as multi-step sequential actions. However, if the \textbf{\textit{Example}} is removed, the agent might skip calling \texttt{autoDownloadOpenStreetMapFile}, resulting in an incorrect answer.  

There is no clear influence from each component's removal on \texttt{queryAreaRange}. This is because there are solely two steps in the task, match the API request and post to the online map server for response, the format of the request body is well-defined, and only execution is needed, which makes it a simple procedure, not sensitive to the prompt component.

\begin{table}
\centering
\caption{Examples of correct and wrong API usage of Augmented Language Model.}
\resizebox{1.0\textwidth}{!}{
\begin{tabular}{|c|c|c|c|}
\hline
\textbf{Error Name}           & \textbf{Question}  & \textbf{Tool} & \textbf{Analysis}   \\ \hline
\multirow{2}{*}{\begin{tabular}[c]{@{}c@{}}\\\\No API Call\\\end{tabular}} & \multirow{2}{*}{\begin{tabular}[c]{@{}c@{}}Can you assist \\ me to download \\the OSM file for \\the Sydney \\Harbour Bridge in Australia?\newline\end{tabular}} & \begin{tabular}[c]{@{}c@{}}\\\textcolor{green}{\ding{51}} : It shows the path to the OSM file\\ for Sydney Harbour Bridge.\\\quad\end{tabular}       & \multirow{2}{*}{\begin{tabular}[c]{@{}c@{}}\\\\It finds a similarly named \\OSM file and shows the wrong answer.\\\end{tabular}} \\ \cline{3-3}
                              &                    & \begin{tabular}[c]{@{}c@{}}\\\textcolor{red}{\ding{55}} : It shows the path to the OSM file\\ for the State of Liberty in New York.\\\\\end{tabular}       &                     \\ \hline

\multirow{2}{*}{\begin{tabular}[c]{@{}c@{}}\\\\API Mismatch\\\quad\end{tabular}} & \multirow{2}{*}{\begin{tabular}[c]{@{}c@{}}I'm interested\\ in the OSM file\\ for Dubai Mall;\\ Can you guide\\ me on that?\\\quad\end{tabular}} & \begin{tabular}[c]{@{}c@{}}\\\textcolor{green}{\ding{51}} : It shows the path to the OSM file\\ for Dubai Mall.\\\quad\end{tabular}       & \multirow{2}{*}{\begin{tabular}[c]{@{}c@{}}\\The question should match\\ \texttt{autoDownloadOpenStreetMapFile} \\but it mismatches with \texttt{queryAreaRange}.\\\quad\end{tabular}} \\ \cline{3-3}
                              &                    & \begin{tabular}[c]{@{}c@{}}\\\textcolor{red}{\ding{55}} : The longitude and latitude of Dubai\\ Mall is [55.274, 25.194, 55.282, 25.199].\\\quad\end{tabular}       &                     \\ \hline

\multirow{2}{*}{\begin{tabular}[c]{@{}c@{}}\\\\

Error Raise\\\quad\end{tabular}}  & \multirow{2}{*}{\begin{tabular}[c]{@{}c@{}}Can you provide\\ the OSM file for\\ CN Tower in\\ Toronto?\\\end{tabular}} & \begin{tabular}[c]{@{}c@{}}\\\textcolor{green}{\ding{51}} : It shows the path to the OSM file\\ for CN Tower in Toronto.\\\quad\end{tabular}       & \multirow{2}{*}{\begin{tabular}[c]{@{}c@{}}\\It mistakenly inputs wrong\\ information into the \texttt{showOnMap}.\\\quad\end{tabular}} \\ \cline{3-3}
                              &                    & \begin{tabular}[c]{@{}c@{}}\\\textcolor{red}{\ding{55}} : Error raise and not keep executing.\\\quad\end{tabular}      &                     \\ \hline
\end{tabular}}
\end{table}

In conclusion, all components contribute to enhancing the performance of \ours. Particularly, the impact of \textbf{\textit{Format Restriction}} and examples is notably significant. This reveals that careful attention should be paid to \textbf{\textit{Format Restriction}} and examples when seeking to enhance the execution-related prompt.

\subsection{Meta Agent Control Analysis}

\paragraph{Experiment Design}
In this section, we conduct experiments to verify the effectiveness of the meta agent control, at the same time, we realized 4 versions of ChatZero on the most well-known LLMs which are: Llama2-7b, Llama2-13b, ChatGPT (GPT-3.5) and GPT-4.0. During the test, the \ours pivotal agent will ask the 4 versions of ChatZero to perform traffic signal control tasks across 4 different traffic configurations using the realistic road map data in Hangzhou city. Each traffic control task is conducted 5 times and the reported results are the $\textit{mean}$ values, following the literature in TSC~\cite{wei2021recent}, the evaluation metrics are as follows: \\
$\bullet$ \textit{Average Travel Time (ATT)} is the average time $t$ it takes for a vehicle to travel through a specific section of a road network. For a control policy, the smaller $ATT$, the better. \\
$\bullet$ \textit{Throughput (TP)} is the number of vehicles that reached their destinations given amount of time. The larger $TP$, the better.\\
$\bullet$ \textit{Reward} is an RL term that measures the return by taking action $a_t$ under state $s_t$. We use the total number of waiting vehicles as the reward, aligned with Preliminaries. The larger the reward, the fewer waiting vehicles, the better.\\
$\bullet$ \textit{Queue} is the number of vehicles waiting to pass through a certain intersection in the road network. Smaller is better.\\
$\bullet$ \textit{Delay} is the average delay per vehicle in seconds and measures the amount of time that a vehicle spends waiting in the network. Smaller is better.

\begin{figure}[h!]
    \centering
    \begin{tabular}{cc}
    \includegraphics[width=0.310\textwidth]     
    {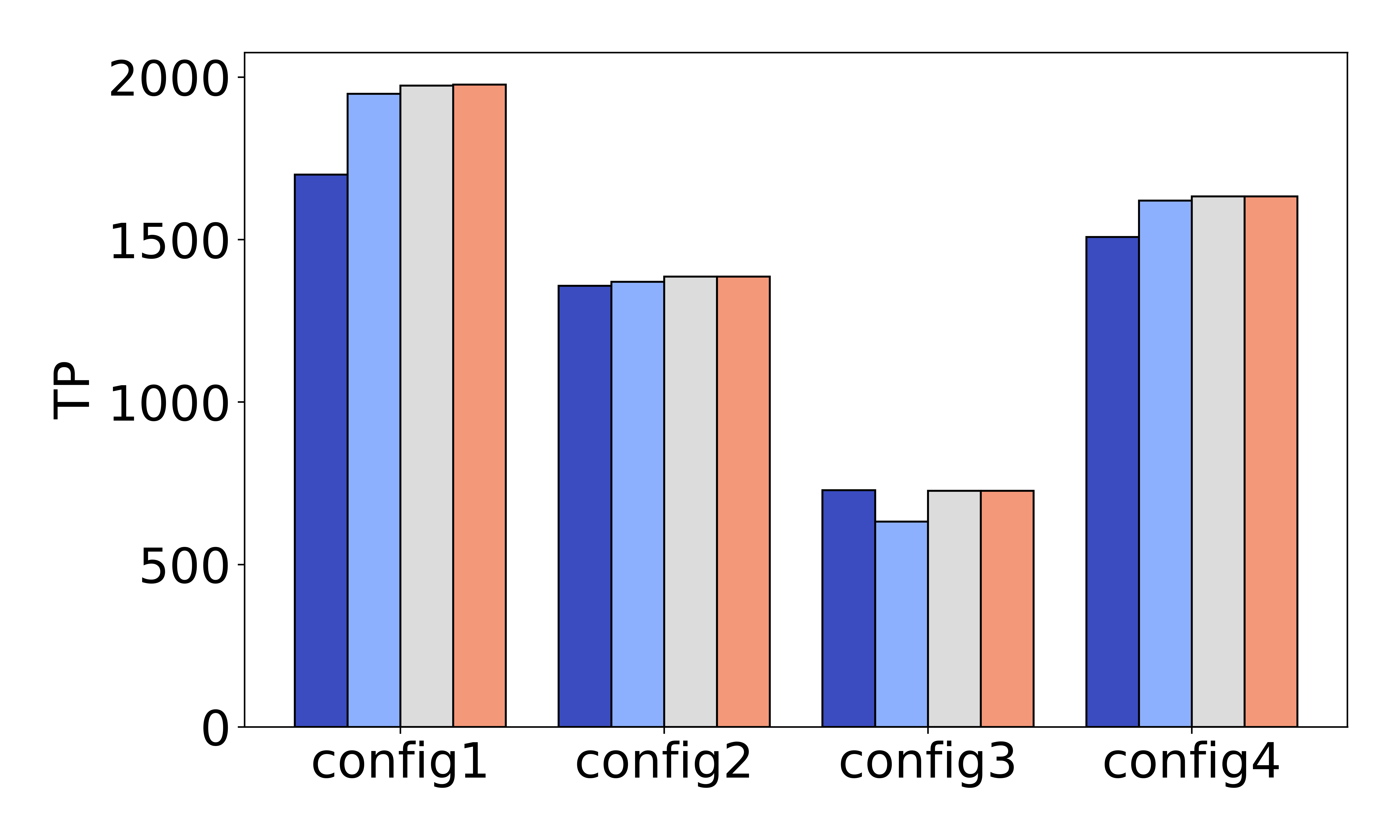}
    &\includegraphics[width=0.310\textwidth]
    {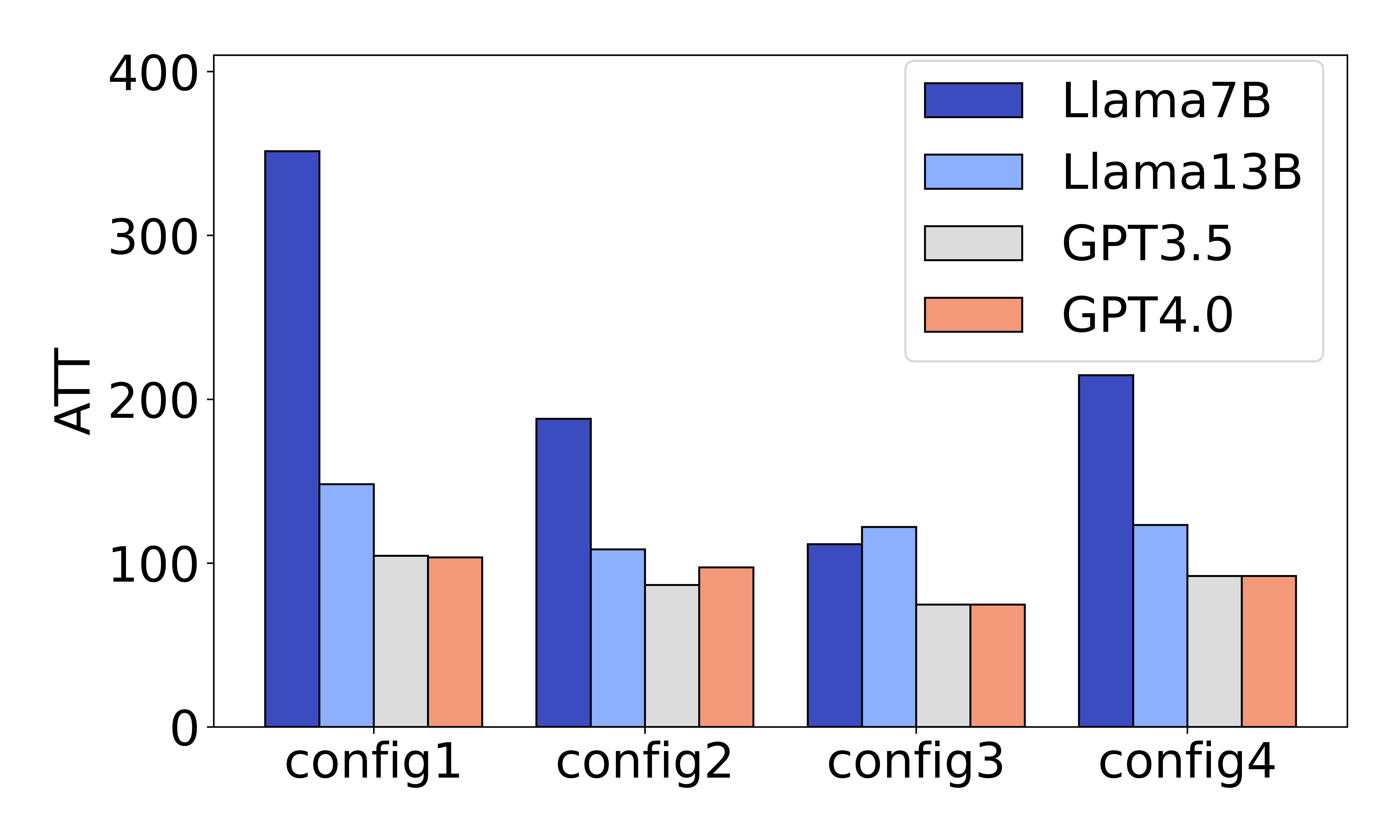} \\
    (a) Throughput & (b) Aevrage Travel Time
    \end{tabular}
    
    \caption{The ChatZero performance in TP and ATT across the 4 LLMs. Each LLM is tested to control the traffic signal in 4 different configs of road situation}
    \label{fig:chatzero}
\end{figure}

\begin{figure}[h!]
    \centering
    \begin{tabular}{ccc}
    \includegraphics[width=0.310\textwidth]     
    {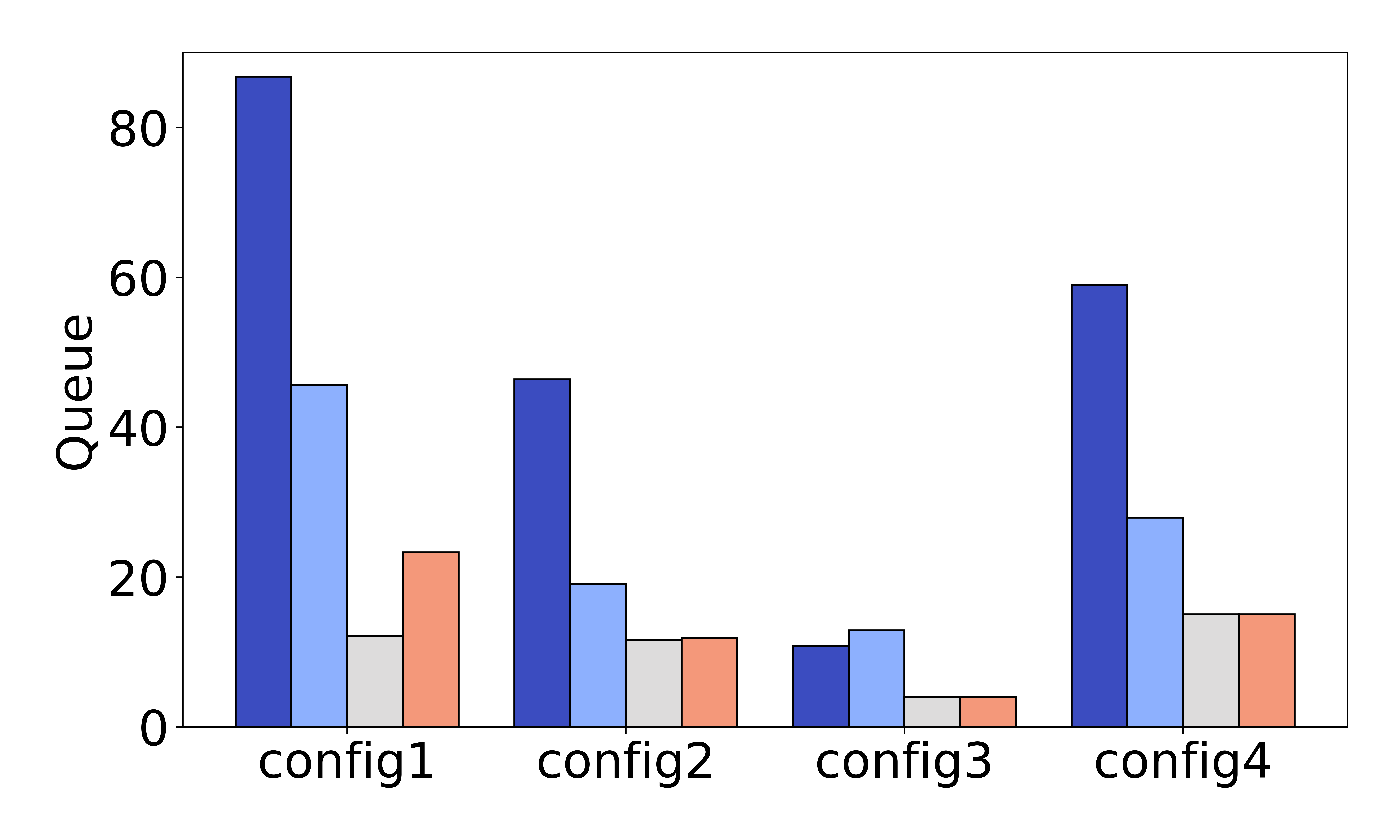}
    &\includegraphics[width=0.310\textwidth]
    {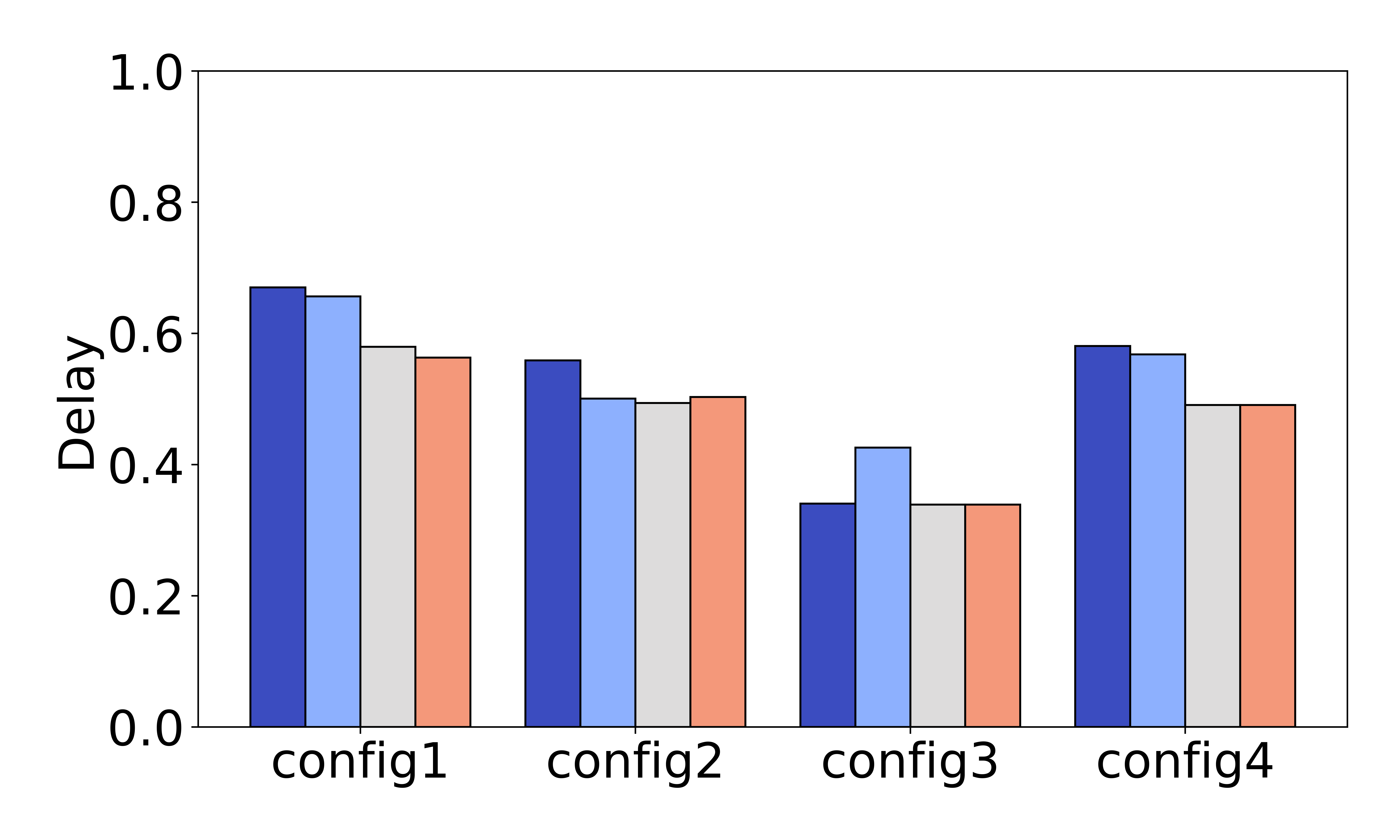}
    & \includegraphics[width=0.310\textwidth]     
    {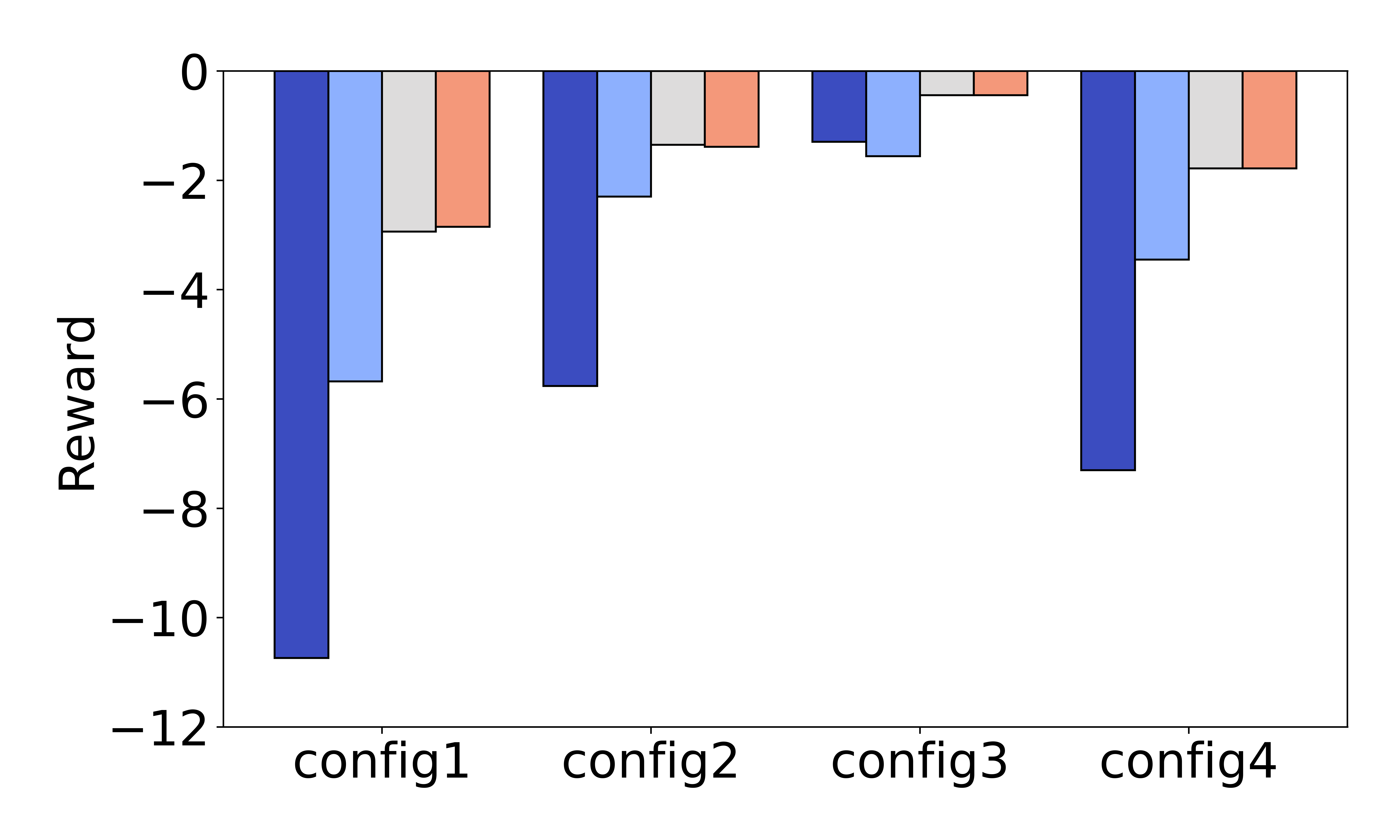}
    \\
    (a) Queue Length & (b) Aevrage Delay  & (c) Reward
    \end{tabular}
    \caption{The ChatZero performance in TP and ATT across the 4 LLMs. Each LLM is tested to control the traffic signal in 4 different configs of road situation}
    \label{fig:chatzero2}
\end{figure}

\paragraph{Experimental Results}
The experiment results are shown in the Fig.~\ref{fig:chatzero}. From the throughput (TP) and average travel time (ATT), we could find out that ChatZero by GPT 4.0 provides the best control results with the overall highest TP and lowest ATT, reflecting its superior ability to understand the policy description and conduct proper control. Other metrics evaluation are shown in Fig.~\ref{fig:chatzero2}.

\section{Conclusion}
In this paper, we propose \ours, an intelligent traffic analysis agent leveraging the large language models' contextual abilities and augmenting them with traffic domain-specific tools, which could provide more than questions consult, but also actual practice on processing raw map data, executing the simulation, training traffic light control policies, and conducting demand optimization, etc. We also explored the meta-control of traffic light by an agent-agent communication scheme named ChatZero,  which could cast insight into the self-explainable control practice. 

We also found that in sequential actions practice, it is easier to occur the API mismatch, for future work, it is important to focus on improving the accuracy of the multi-step action embodiment.  We have provided the code base and an explicit implementation structure for the research community to enhance and expand the \ours's intelligence.

\bibliography{sn-bibliography}

\clearpage
\begin{appendices}

\section{}

\subsection{Thought Chain Process Examples}\label{app:example1}
In this section, we provide more Chain-of-thought (CoT) process examples, as a reflection on given a task, how the \ours thinks and proposes the solutions, and how it searches in the augmentation tools to further provide analysis.

We have shown the requests such as: downloading OSM files of specific locations, interpreting the log files, showing areas on a map, filtering assigned lane types from a given map, generating demand files based on a map file, executing multiple simulations like DLSim, SUMO, etc., running LibSignal for traffic signal control. 

\begin{figure}[h!]
    \centering
    \includegraphics[width=0.95\linewidth]{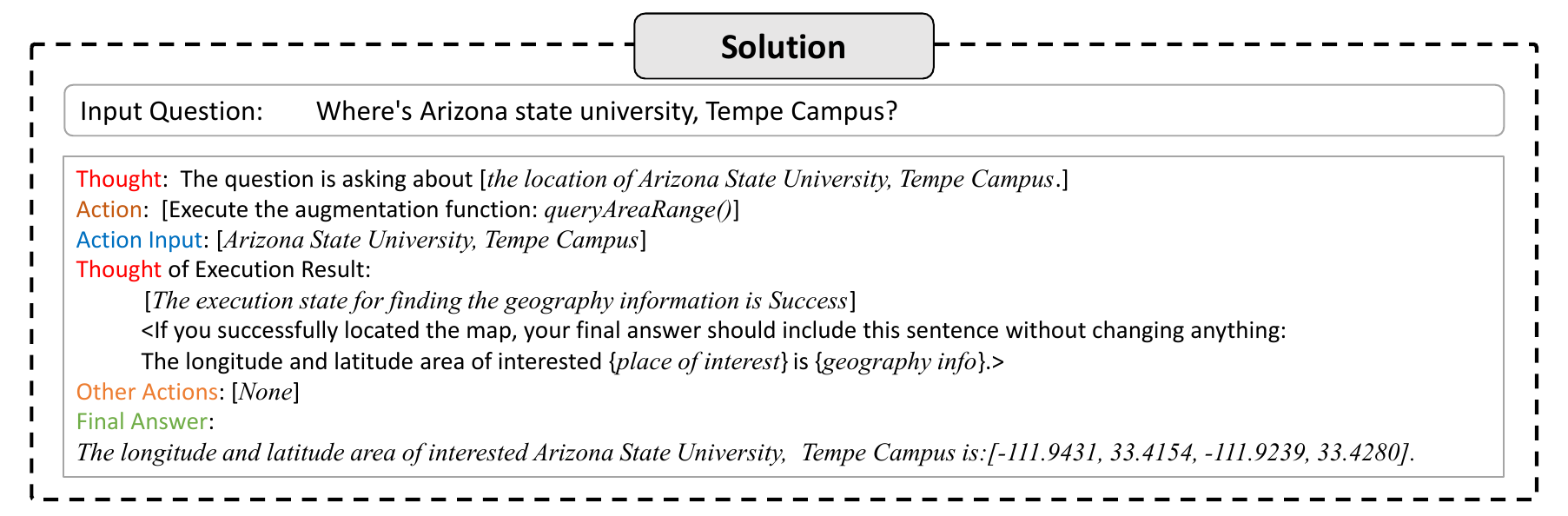}
    \caption{Ask \ours to get the geographic information of Arizona State University, Tempe Campus.}
    \label{fig:process2}
\end{figure}
\begin{figure}[h!]
    \centering
    \includegraphics[width=0.95\linewidth]{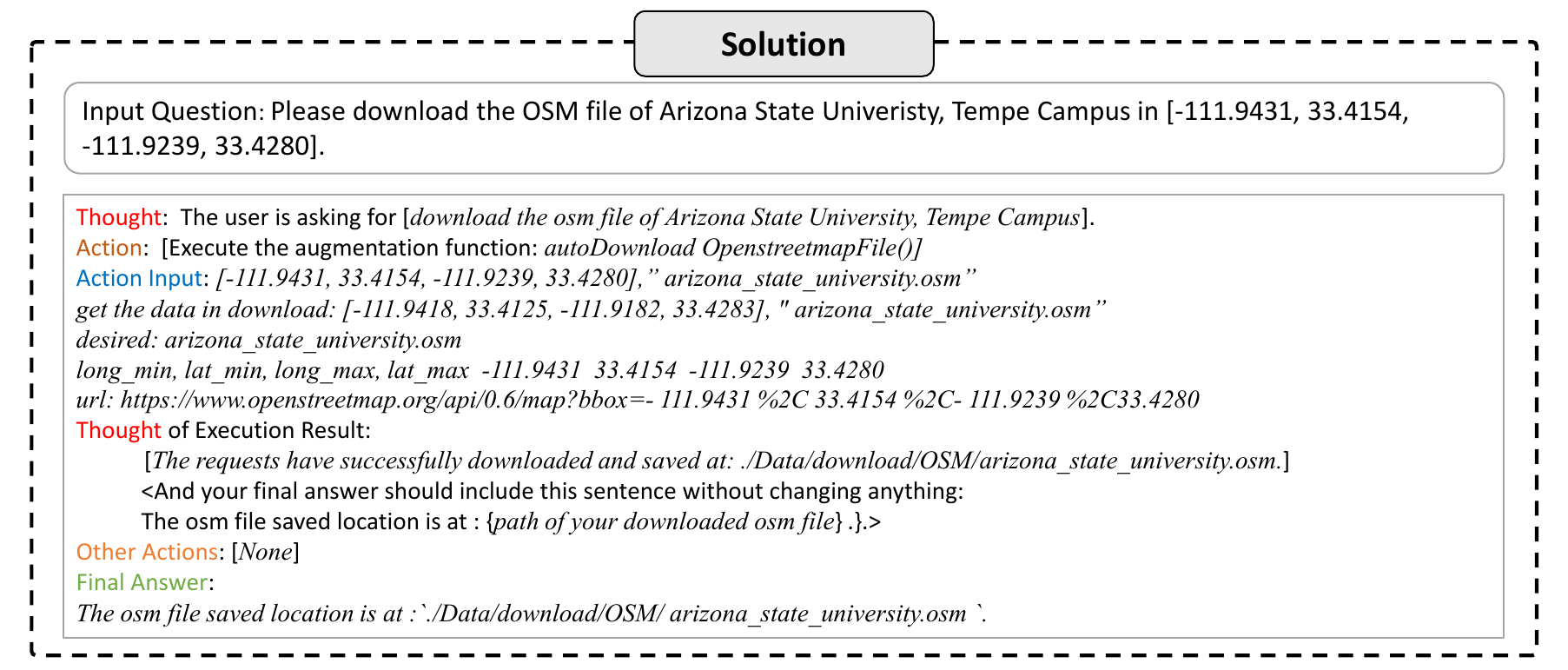}
    \caption{Ask \ours to download OSM data of Arizona State University, Tempe Campus.}
    \label{fig:process2}
\end{figure}
\begin{figure}[h!]
    \centering
    \includegraphics[width=0.95\linewidth]{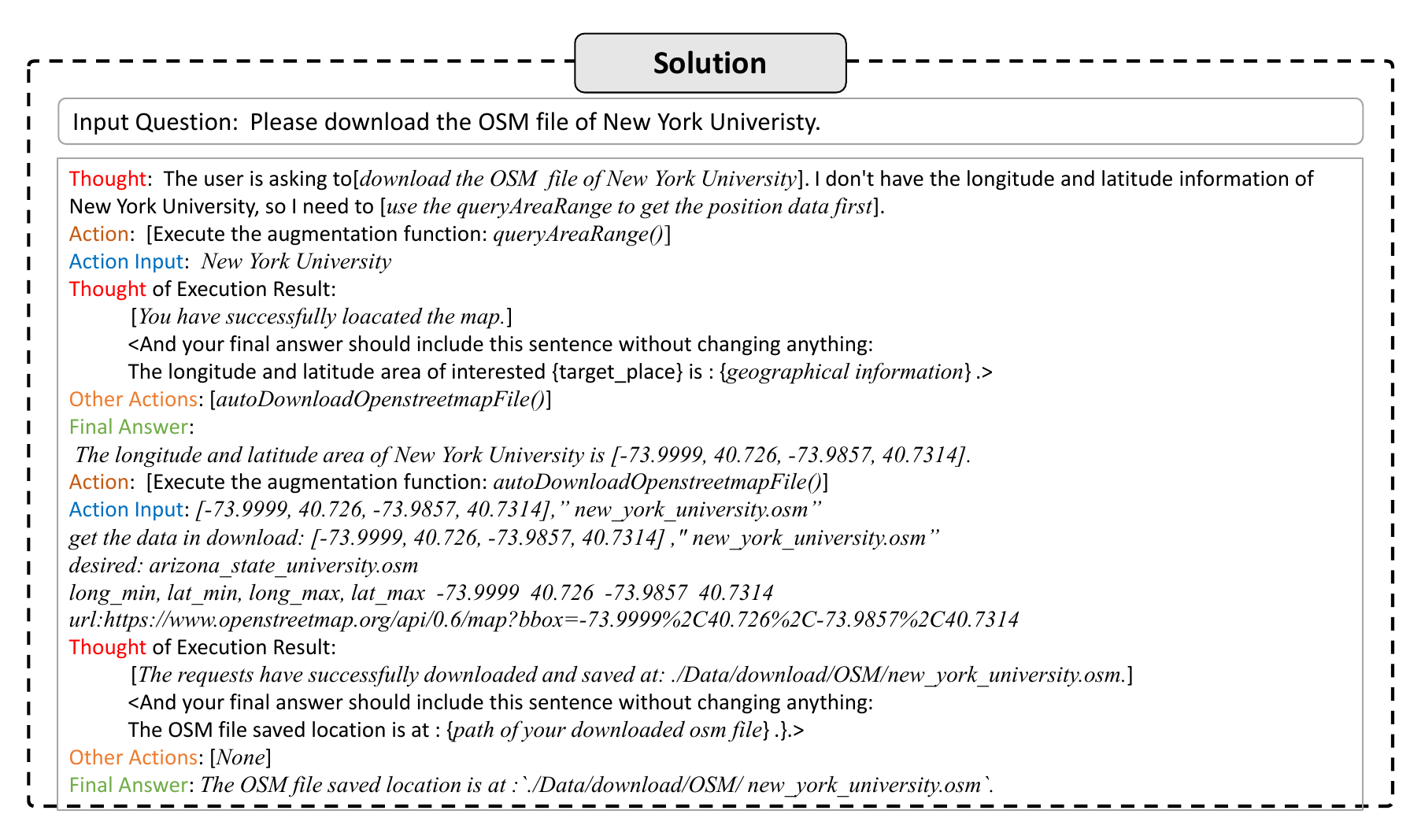}
    \caption{Ask \ours to download OSM data of Arizona State University, Tempe Campus without giving \ours geographic information.}
    \label{fig:process2}
\end{figure}
\begin{figure}[h!]
    \centering
    \includegraphics[width=0.95\linewidth]{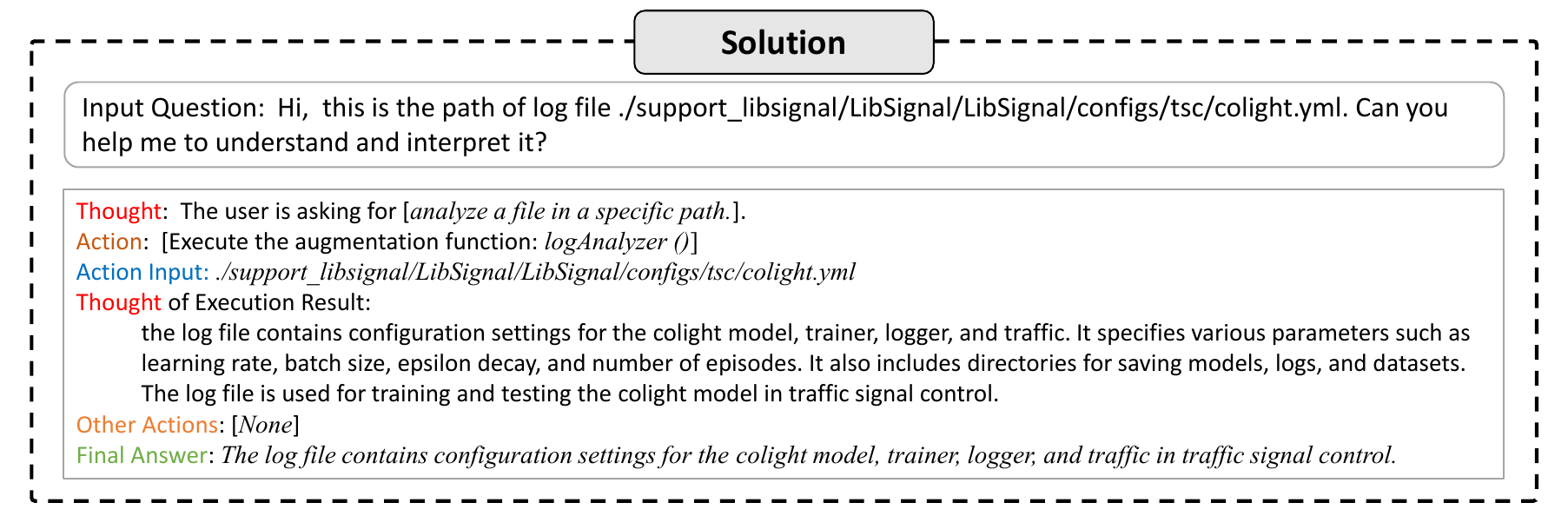}
    \caption{Ask \ours to analyze the interested log file in a specific path. }
    \label{fig:process2}
\end{figure}
\begin{figure}[h!]
    \centering
    \includegraphics[width=0.95\linewidth]{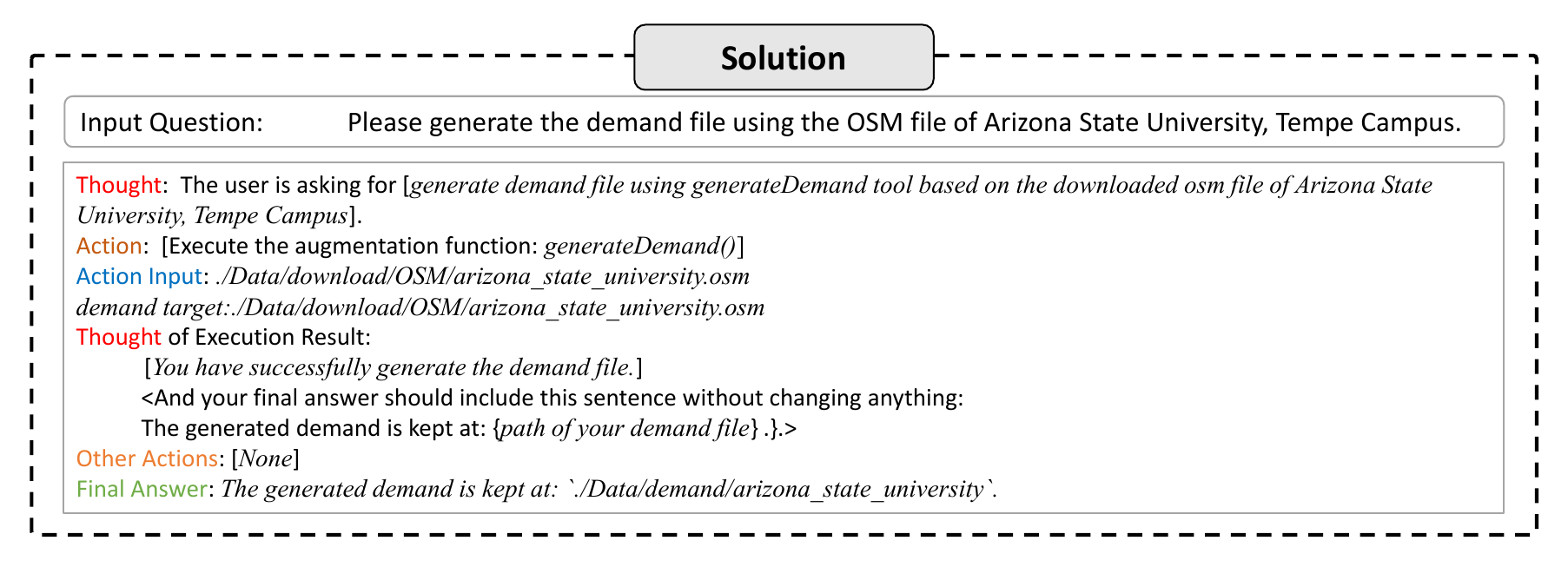}
    \caption{Ask \ours to generate demand file through downloaded osm data of Arizona State University, Tempe Campus.}
    \label{fig:process2}
\end{figure}

\begin{figure}[t!]
    \centering
    \includegraphics[width=0.95\linewidth]{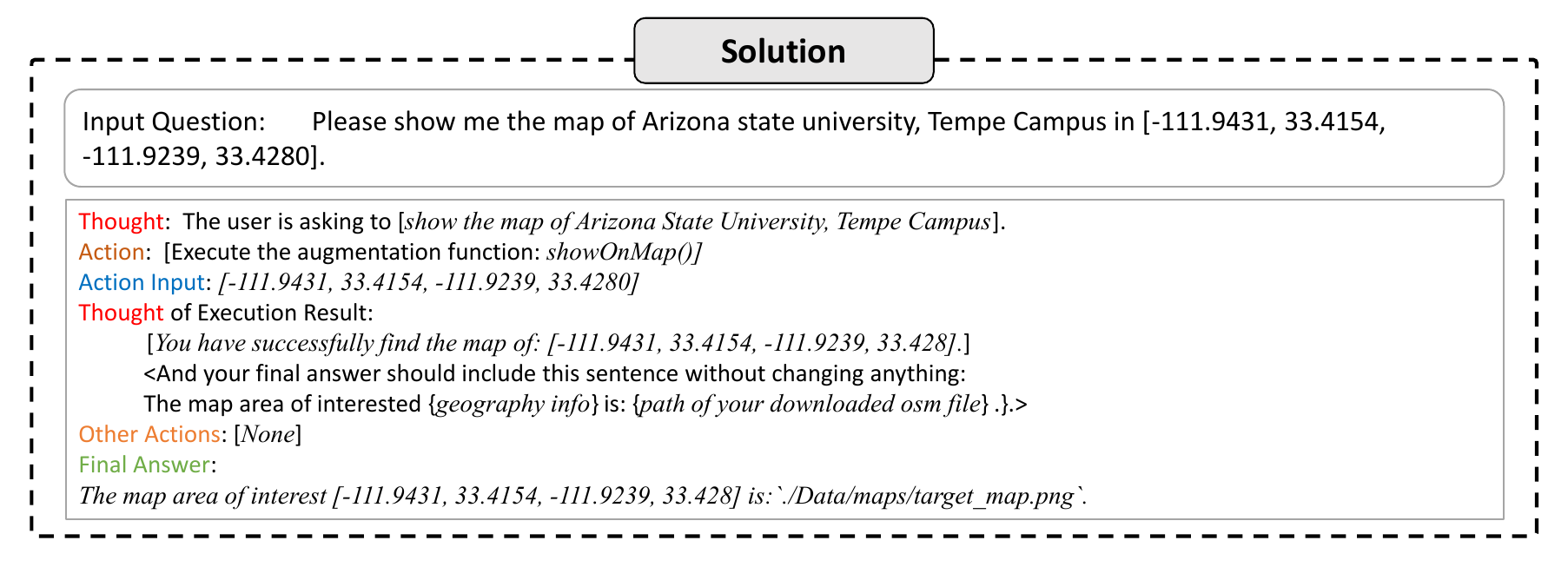}
    \caption{Ask \ours to show the map of Arizona State University, Tempe Campus.}
    \label{fig:process2}
\end{figure}

\begin{figure}[h!]
    \centering
    \includegraphics[width=0.95\linewidth]{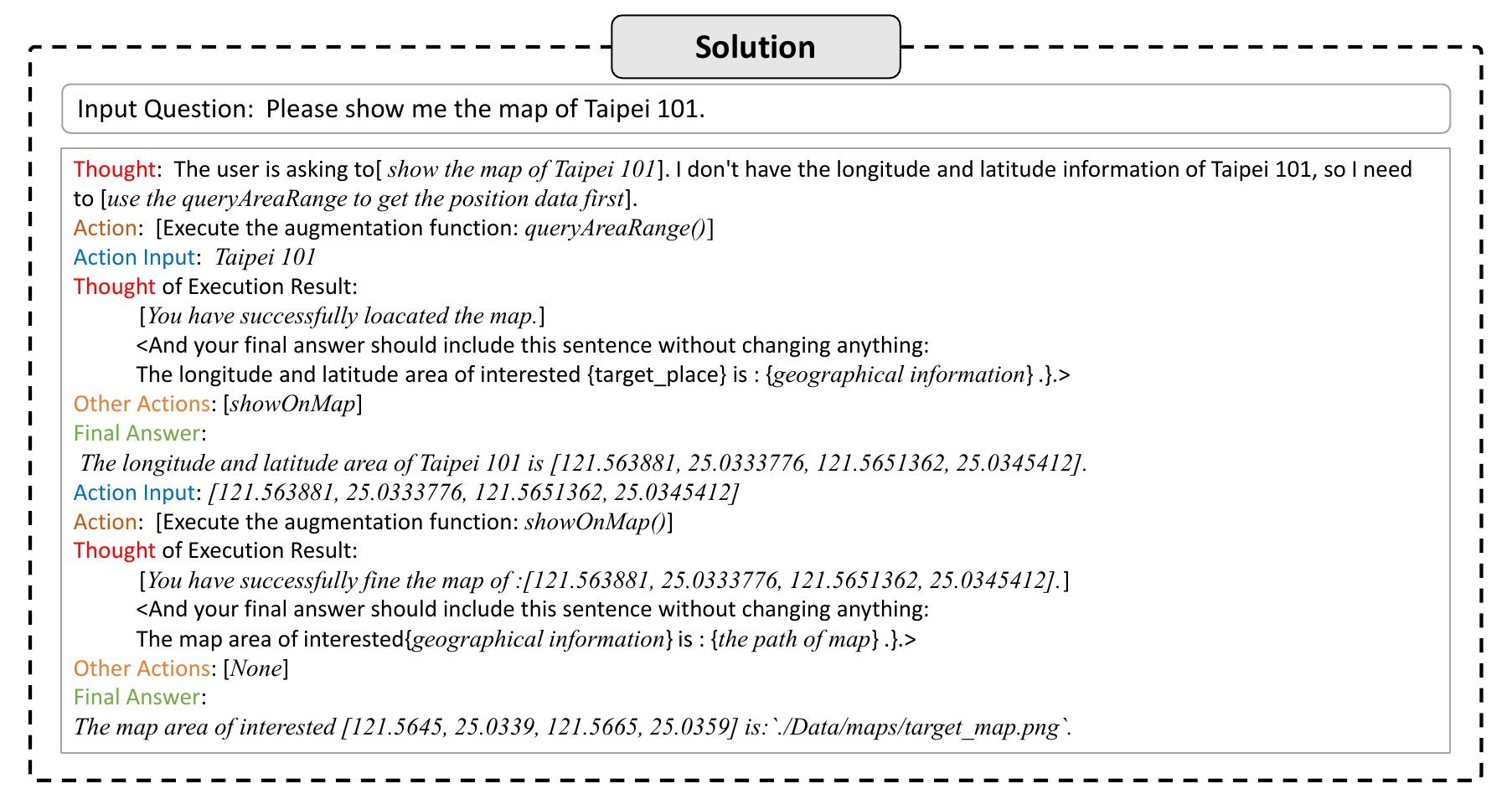}
    \caption{Ask \ours to show the map of Taipei 101 without giving \ours geographic information.}
    \label{fig:process2}
\end{figure}
\begin{figure}[h!]
    \centering
    \includegraphics[width=0.95\linewidth]{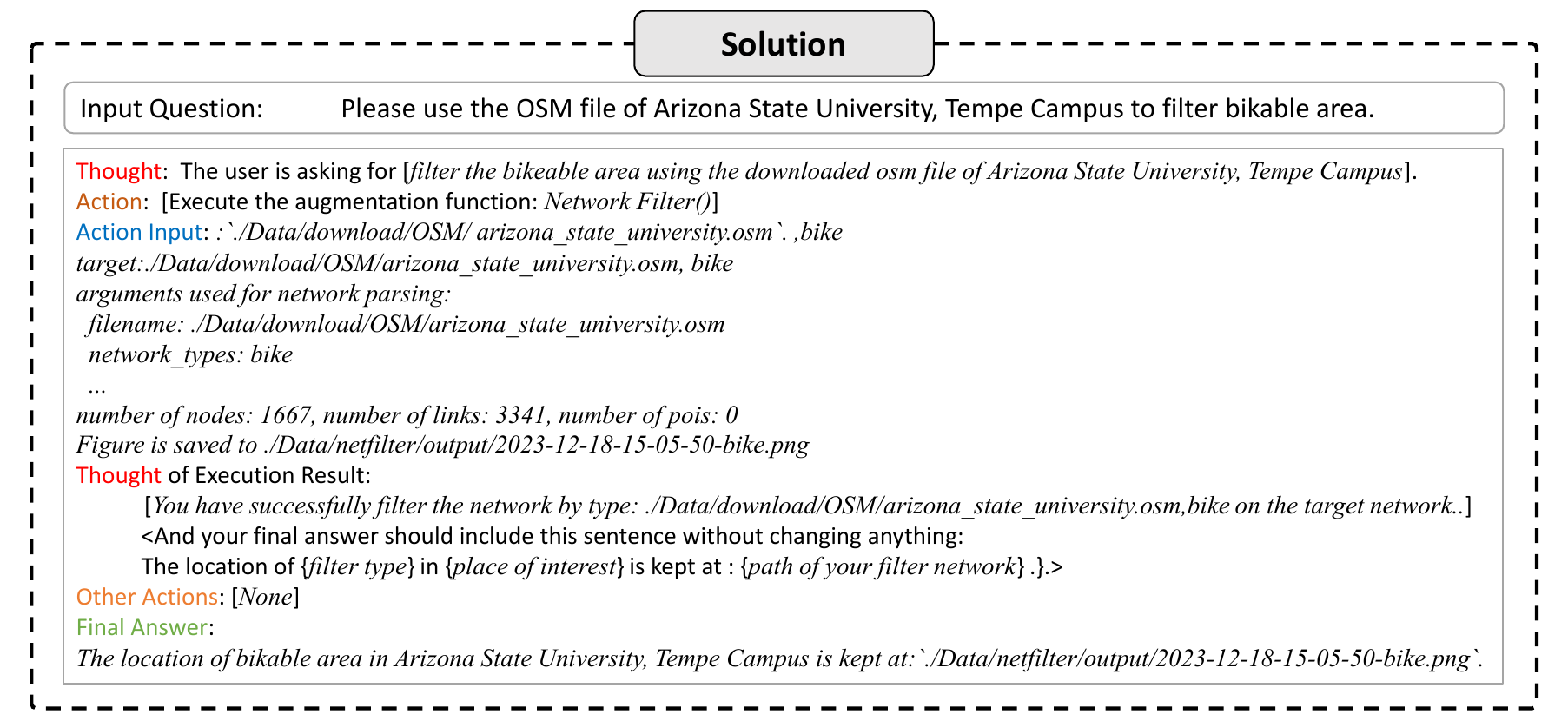}
    \caption{Ask \ours to filter the bikable area  of Arizona State University, Tempe Campus.}
    \label{fig:process2}
\end{figure}

\begin{figure}[h!]
    \centering
    \includegraphics[width=0.95\linewidth]{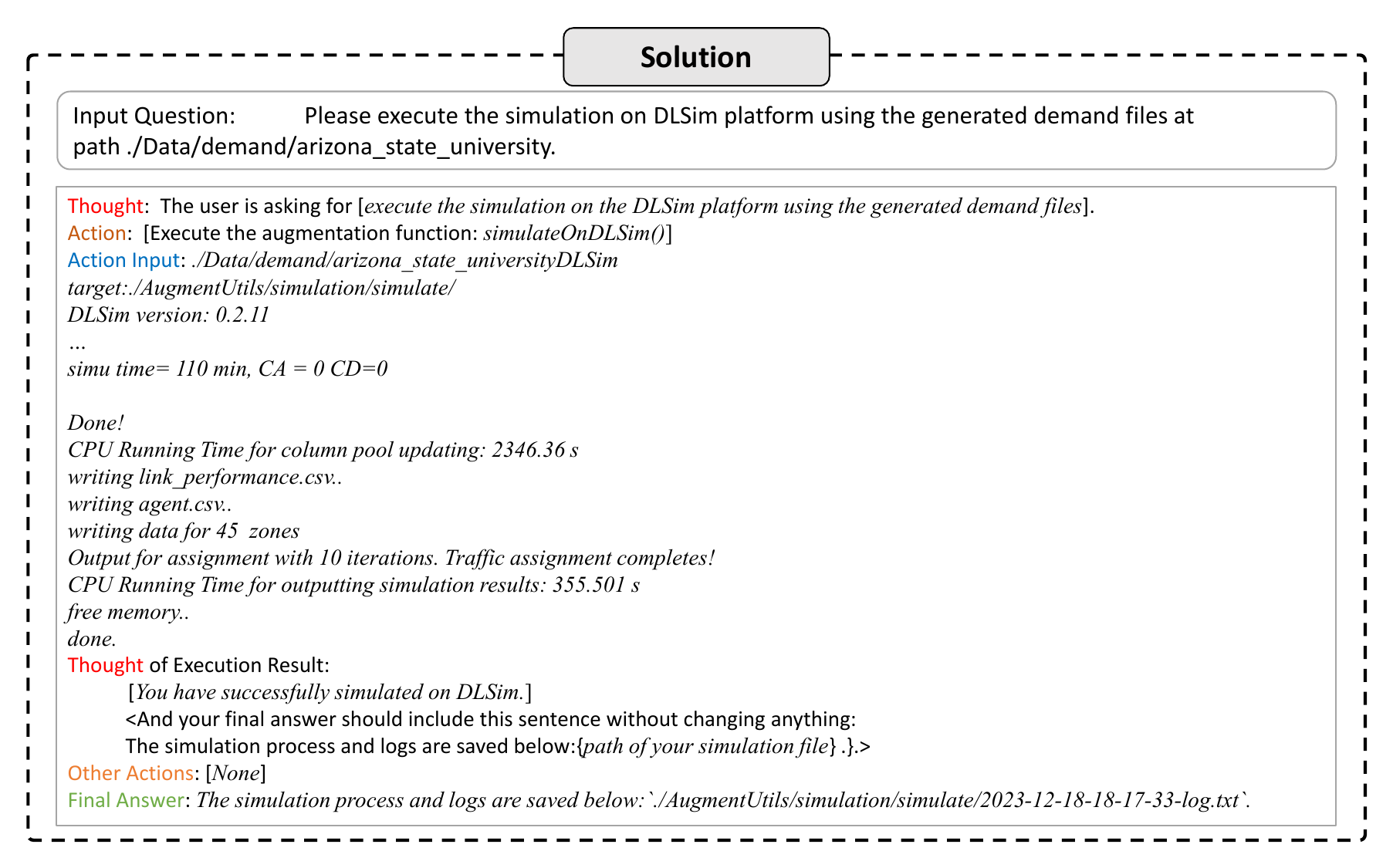}
    \caption{Ask \ours to run the demand file of Arizona State University, Tempe Campus on DLSim. }
    \label{fig:process2}
\end{figure}
\begin{figure}[h!]
    \centering
    \includegraphics[width=0.95\linewidth]{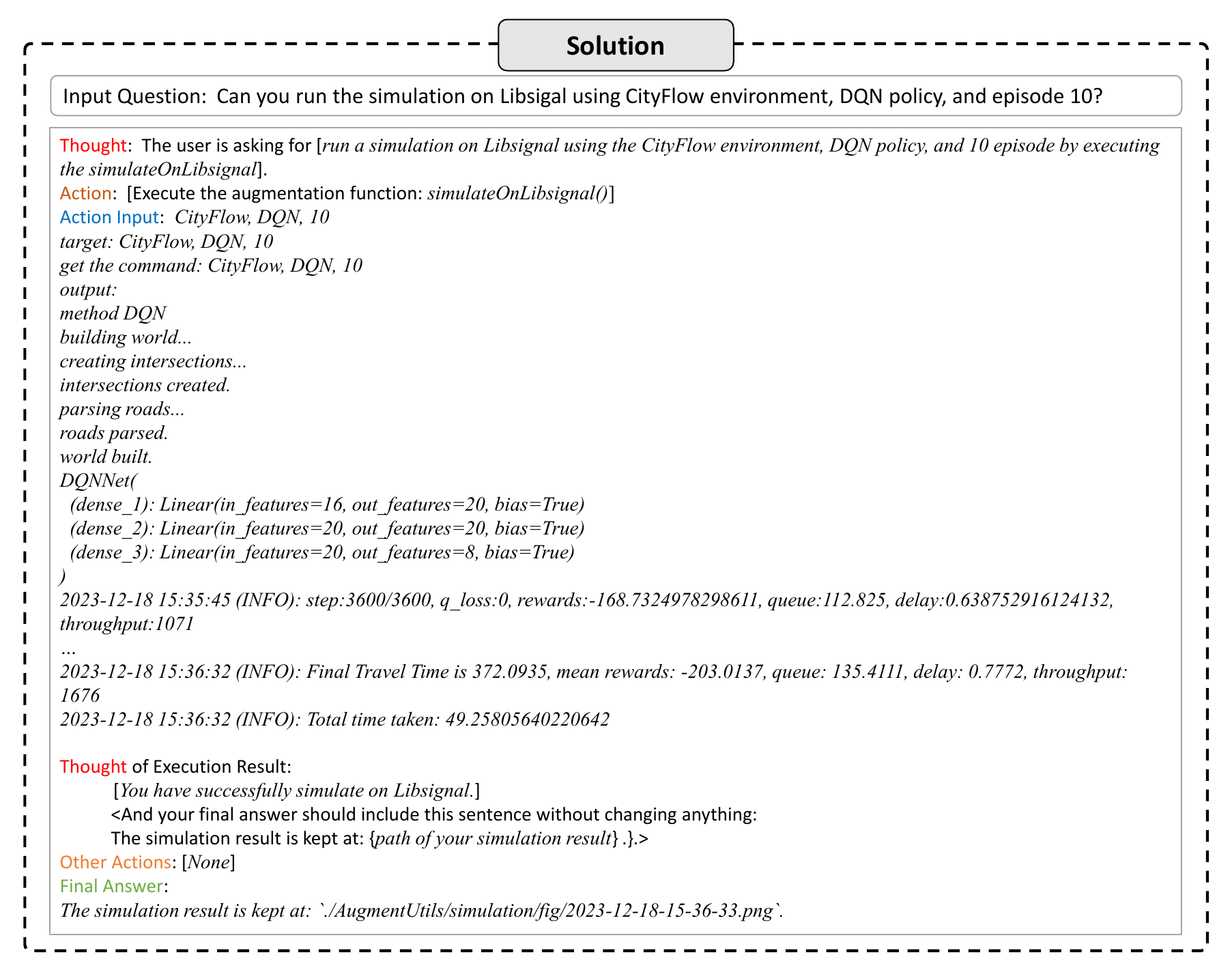}
    \caption{Ask \ours to run Libsignal on CityFlow environment, DQN policy, and episode 10.}
    \label{fig:process2}
\end{figure}

\clearpage
\subsection{Other interactions with \ours examples}\label{app:example2}
This section provides more examples of user interactions, including result interpretation, log file analysis, O-D matrix optimization, etc.

\begin{figure}[h!]
    \centering
    \includegraphics[width=0.95\linewidth]{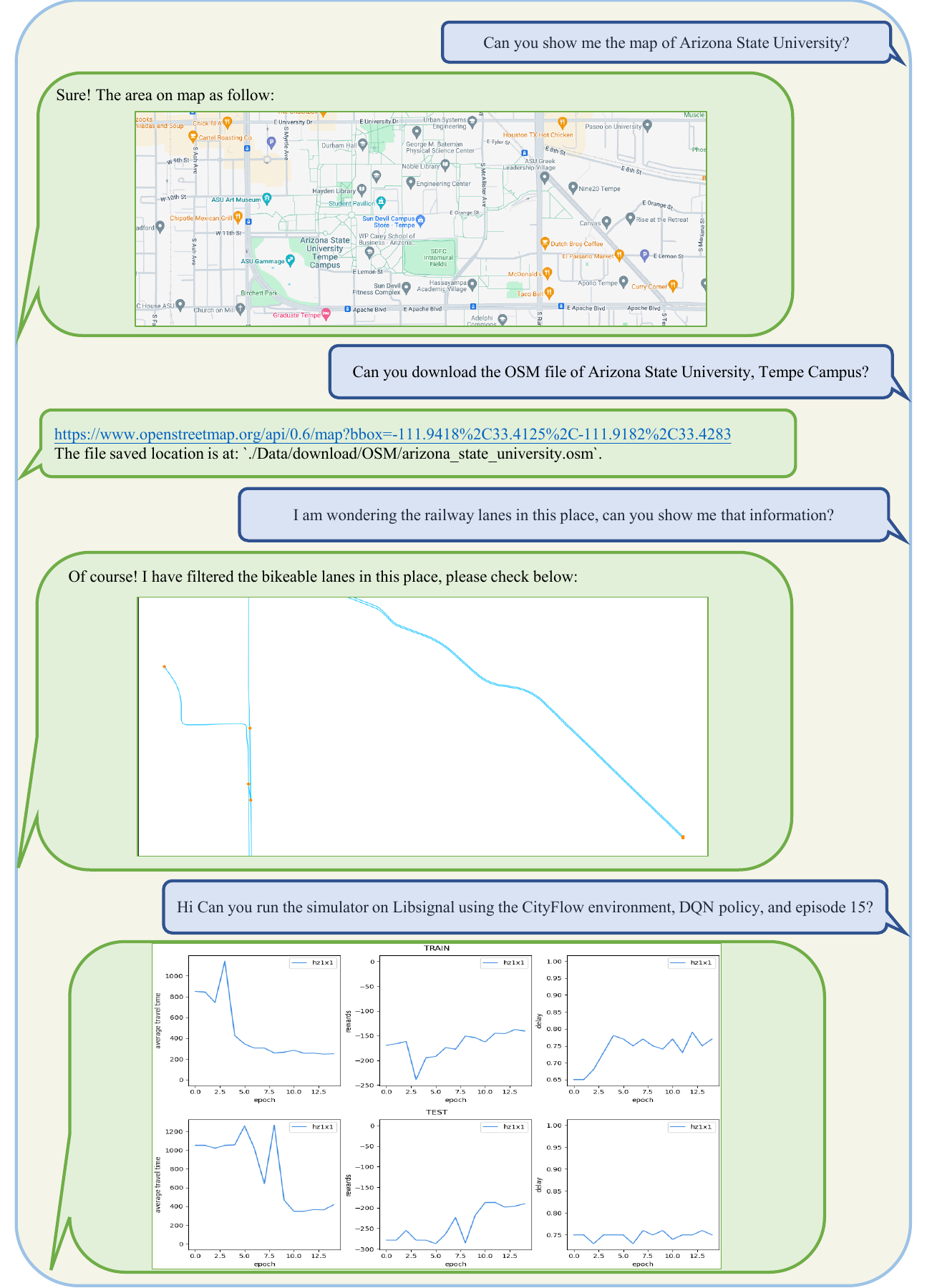}
    \caption{Ask \ours to show the map of interested place,  download .osm data of interested place, use the OSM file of target place to filter railway routes, and conduct traffic signal control.}
    \label{fig:process2}
\end{figure}

\begin{figure}[h!]
    \centering
    \includegraphics[width=0.95\linewidth]{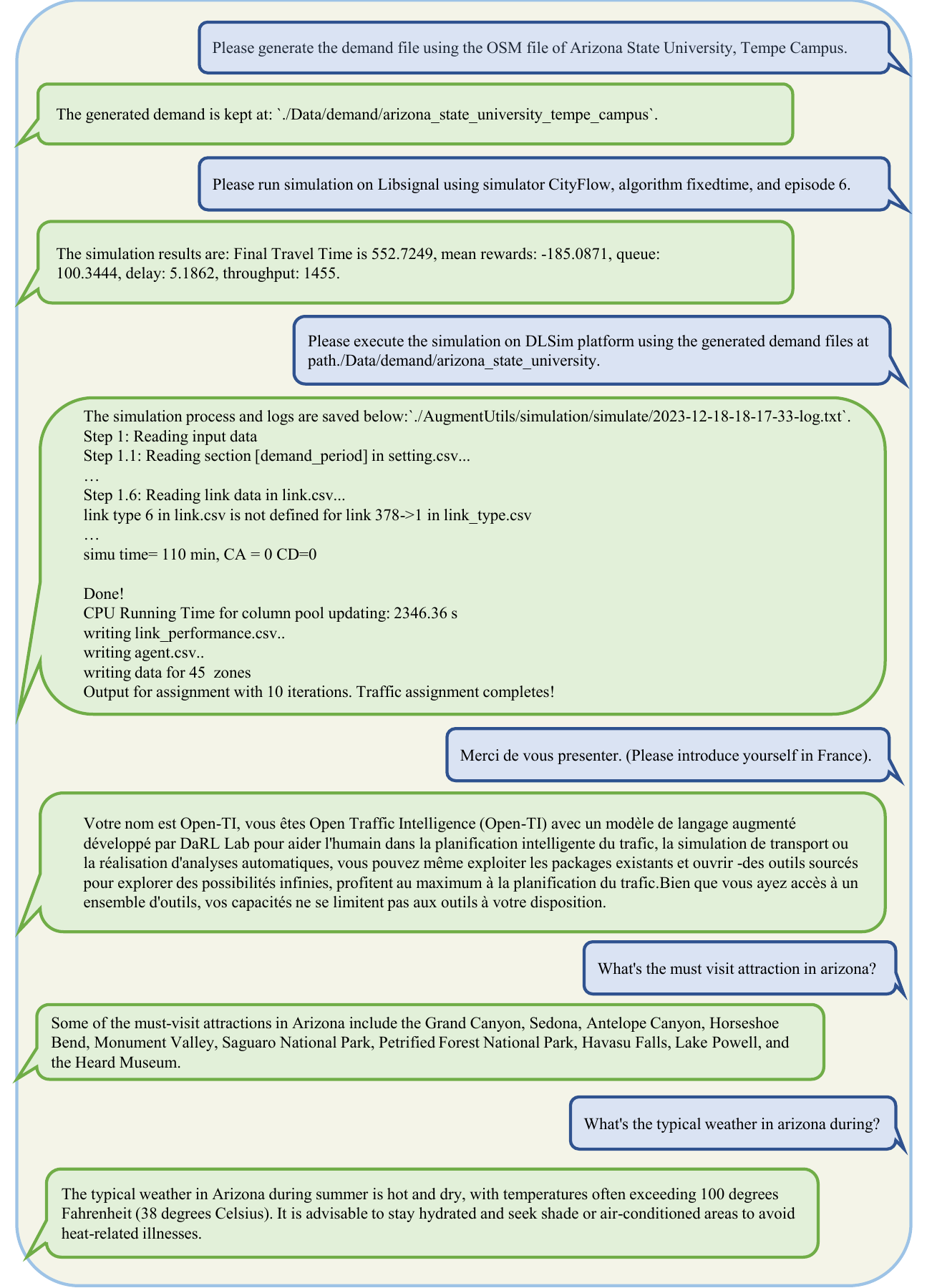}
    \caption{An example of asking \ours to generate demand file from OSM file, run simulator on Libsignal in different algorithm and episode, execute simulation on DLSim, understand and response in multi-languages, and analyze logs in a specific path.}
    \label{fig:process2}
\end{figure}

\begin{figure}[h!]
    \centering
    \includegraphics[width=0.95\linewidth]{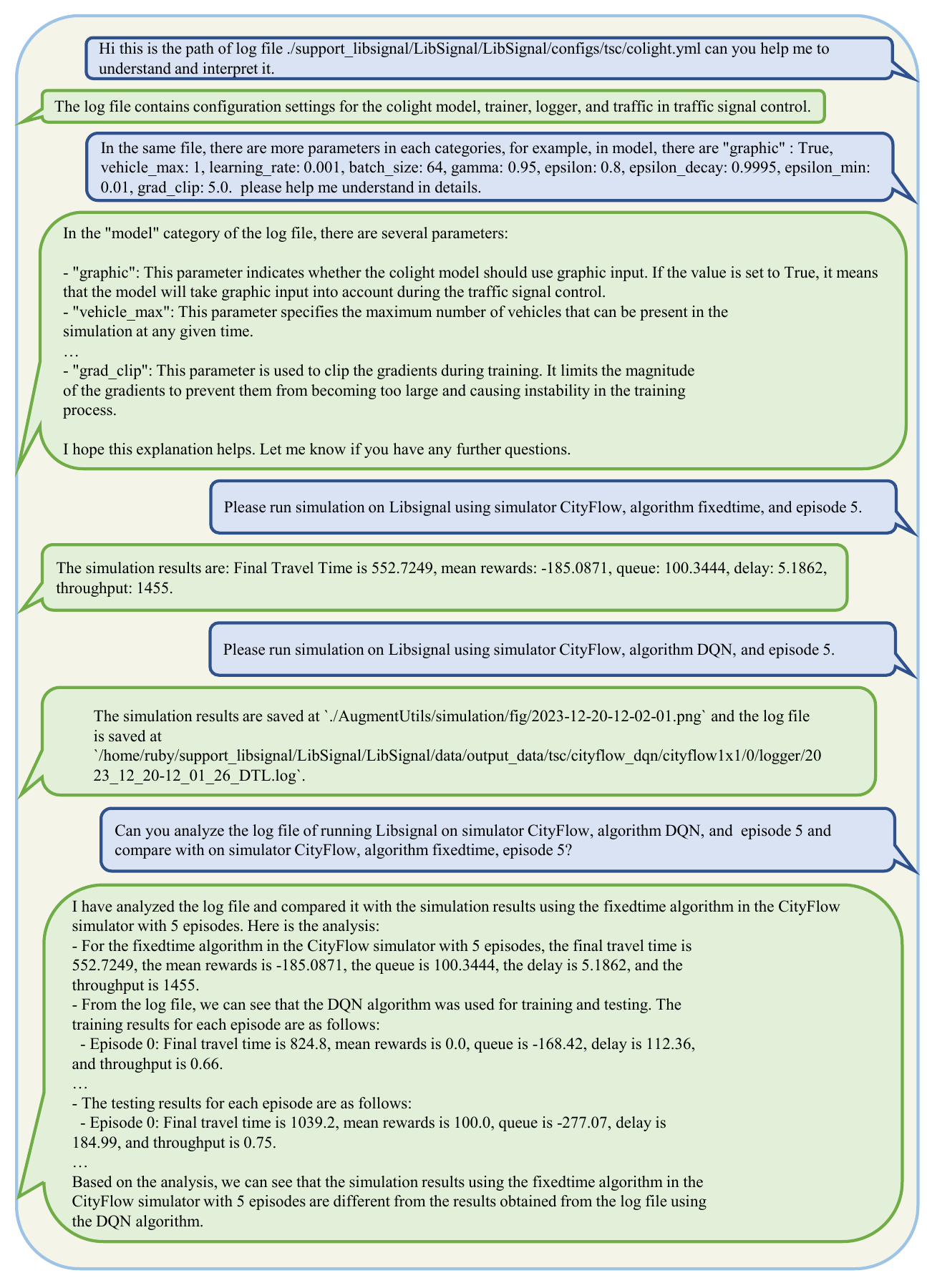}
    \caption{An example of asking \ours to analyze file with a specific path. In this case,  run the simulation on Libsignal using simulator CityFlow, algorithm fixedtime, and episode 5. Additionally,  run the simulation on Libsignal using simulator CityFlow, algorithm DQN, and episode 5. Finally, use the \texttt{logAnalyzer} to compare the performance of both algorithms.}
    \label{fig:process2}
\end{figure}

\begin{figure}[h!]
    \centering
    \includegraphics[width=0.95\linewidth]{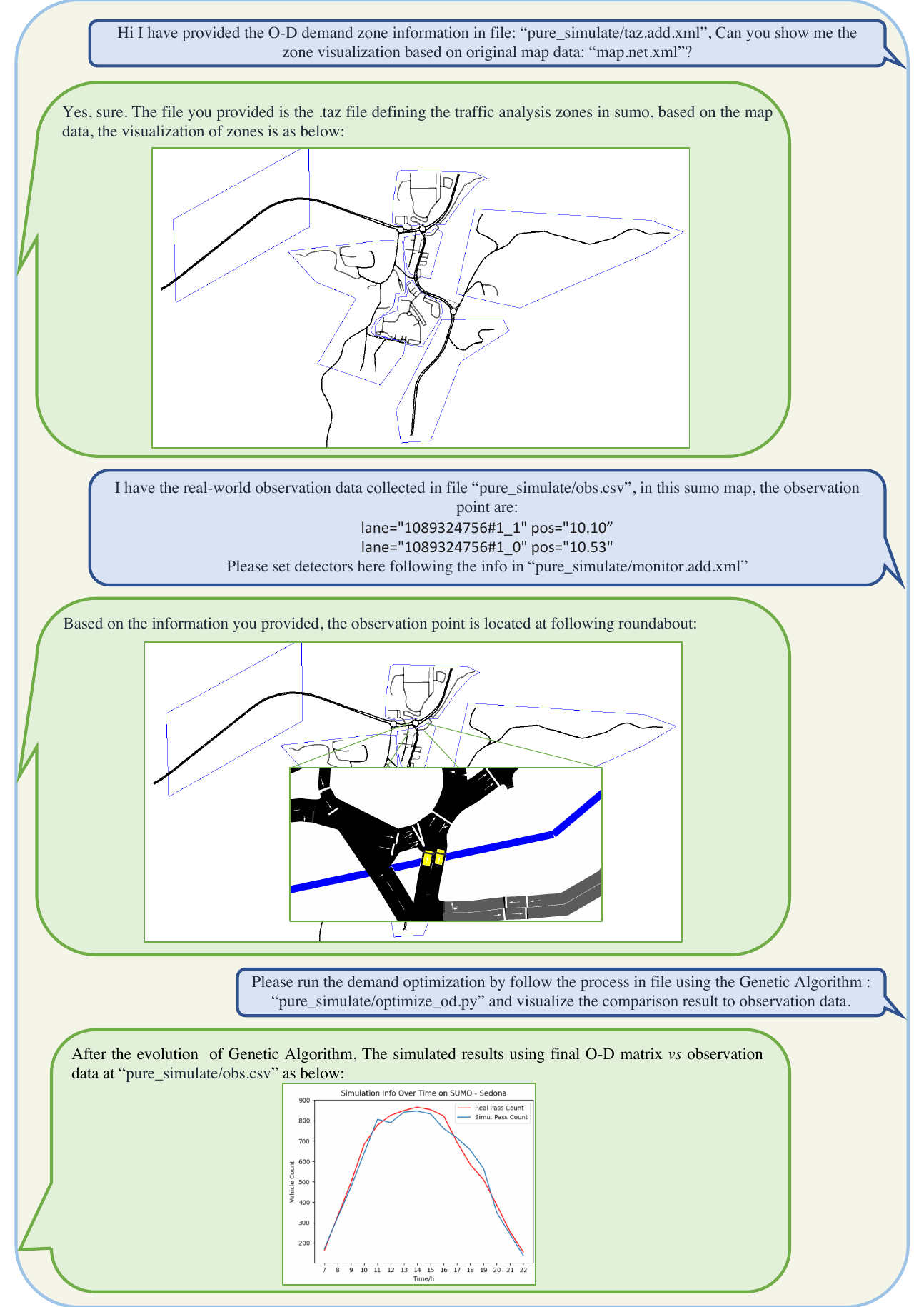}
    \caption{The demonstration on how \ours is used to optimize the OD-Demand matrix. It first visualizes the defined traffic zone information, and sets the observation point to mimic the real-world data collection process. Then based on the gap between simulation observation and real-world observation (count data), the O-D matrix is optimized to mitigate the observation gap by optimization algorithms (e.g., Genetic Algorithm). After the optimization, the final O-D matrix is simulated again, and the comparison of observation is shown in the end.}
    \label{fig:odmatrix}
\end{figure}




\end{appendices}

\end{document}